\documentclass{article}

\PassOptionsToPackage{numbers, compress}{natbib}

\usepackage[preprint]{style}

\usepackage[utf8]{inputenc} 
\usepackage[T1]{fontenc}    
\usepackage{hyperref}       
\usepackage{url}            
\usepackage{booktabs}       
\usepackage{amsfonts}       
\usepackage{nicefrac}       
\usepackage{microtype}      

\usepackage{amsmath}
\usepackage{amssymb}
\usepackage{xcolor}
\usepackage[nolist]{acronym}
\usepackage{placeins}
\usepackage{float}
\usepackage{enumitem}%
\usepackage{graphicx}
\usepackage{subcaption}
\usepackage{algorithmic}
\usepackage{algorithm}

\usepackage{tabulary}
\usepackage[font=small,labelfont=bf,tableposition=top]{caption}

\DeclareCaptionLabelFormat{andtable}{#1~#2  \&  \tablename~\thetable}

\definecolor{dred}{HTML}{9c4444}
\definecolor{blue}{HTML}{417AB8}
\definecolor{yellow}{HTML}{cf990a}
\definecolor{grey}{HTML}{979797}

\newcommand{\rc}[1]{\textcolor{red}{#1}}
\renewcommand{\vec}[1]{\mathbf{#1}}

\begin{acronym}
\acro{RNN}{Recurrent Neural Network}
\acro{GRU}{gated recurrent unit}
\acro{LSTM}{long short-term memory}
\acro{MFCC}{Mel-frequency cepstral coefficient}
\acro{PEM}{per-epoch noise mixing}
\acro{SNR}{signal-to-noise ratio}
\acro{ASR}{automatic speech recognition}
\acro{CNN}{Convolutional Neural Network}
\acro{DNN}{Deep Neural Network}
\acro{ACCAN}{accordion annealing}
\acro{WSJ}{Wall Street Journal}
\acro{CTC}{Connectionist Temporal Classification}
\acro{WFST}{Weighted Finite State Transducer}
\acro{WER}{word error rate}
\acro{CER}{character error rate}
\acro{LVCSR}{large-vocabulary continuous speech recognition}
\acro{ROI}{range of interest}
\acro{STAN}{sensor transformation attention network}
\acro{SER}{sequence error rate}
\acro{HMM}{Hidden Markov Model}
\acro{LER}{label error rate}
\acro{NN}{Neural Networks}
\acro{MLP}{Multilayer Perceptron }
\acro{ReLU}{Rectified Linear Unit}
\acro{PCC}{Pearson correlation coefficient}
\acro{LPN}{Lightweight Probabilistic Deep Network}
\end{acronym}

\title{Shapley Values as a Principled Metric for\\ Structured Network Pruning}

\author{%
  Marco Ancona \\
  Department of Computer Science\\
  ETH Zurich, Switzerland\\
  \texttt{marco.ancona@inf.ethz.ch} \\
   \And
   Cengiz {\"O}ztireli \\
   Department of Computer Science and Technology \\ University of Cambridge, UK \\
   \texttt{aco41@cam.ac.uk} \\
   \AND
   Markus Gross \\
   Department of Computer Science\\
   ETH Zurich, Switzerland\\
   \texttt{grossm@inf.ethz.ch} \\
}

\begin{document}
\maketitle

\begin{abstract}
Structured pruning is a well-known technique to reduce the storage size and inference cost of neural networks.
The usual pruning pipeline consists of ranking the network internal filters and activations with respect to their contributions to the network performance, removing the units with the lowest contribution, and fine-tuning the network to reduce the harm induced by pruning. 
Recent results showed that random pruning performs on par with other metrics, given enough fine-tuning resources.
In this work, we show that this is not true on a low-data regime when fine-tuning is either not possible or not effective. 
In this case, reducing the harm caused by pruning becomes crucial to retain the performance of the network.
First, we analyze the problem of estimating the contribution of hidden units with tools suggested by cooperative game theory and propose Shapley values as a principled ranking metric for this task. 
We compare with several alternatives proposed in the literature and discuss how Shapley values are theoretically preferable.
Finally, we compare all ranking metrics on the challenging scenario of low-data pruning, where we demonstrate how Shapley values outperform other heuristics.
\end{abstract}

\section{Introduction}
\label{introduction}

Pruning, as a method to compress \acp{DNN} and hence to reduce their memory footprint and inference time, has gained significant interest in the research community as well as in the industry over the last decade. Several works have shown that these models are often highly overparameterized and that it is possible to prune a part of the weights, activations, or layers without degrading the network performance significantly \citep{lecun1990optimal, han2015learning}.
When discussing pruning, the literature distinguishes between unstructured and structured pruning. With \textit{unstructured pruning}, the network weights can be removed independently from each other \cite{han2015learning, frankle2018the, lee2018snip, wang2020picking}. While this is very effective for network compression (as it leaves the pruning algorithm many degrees of freedom to operate on the network connectivity), the pruning does not translate directly into a faster inference on modern hardware, which is not optimized for sparse computations. 
In this work, we focus instead on \textit{structured pruning} \cite{hu2016network, li2016pruning, molchanov2016pruning, luo2017thinet, he2017channel}. In this case, the size of the weight tensor is reduced by a slicing operation, thus also immediately reducing the number of floating-point operations (FLOPs) required for the forward and backward passes. For densely connected layers, this translates into removing a neuron, with all its incoming and outgoing edges. For convolutional layers, the most common approach is to remove convolutional filters \cite{hu2016network, molchanov2016pruning, he2017channel, li2016pruning}. 
In the remainder of the paper, we will refer to structured pruning every time we talk about pruning without further specification. 

Typically, pruning is performed by first training a large network and then removing the least important units according to the ranking provided by an attribution metric, sometimes also called ``importance score'' \cite{li2016pruning} or ``pruning criterion'' \cite{molchanov2016pruning, mittal2019studying}. This metric essentially estimates the importance of each unit for the end task, thus allowing the identification of the least relevant units as the ones to be pruned, up to a target sparsity ratio.
Several previous works proposed attribution metrics based on heuristics or linear approximations of the model \cite{li2016pruning,hu2016network,molchanov2016pruning,mittal2019studying}.
Notably, empirical comparisons on image classification and object detection tasks showed that there is no significant difference between these metrics and that they do not yield significantly better results than random pruning \cite{mittal2019studying}.
In this paper, we complement this analysis showing instead theoretical differences as well as significant performance gaps in a low-data regime, when the original training data is not available for fine-tuning, e.g., when the original training data is not public or in a transfer learning setting.

Our contributions are as follows: 1) by analyzing the pruning problem on a simple interpretable model, we suggest a number of desirable properties that an ideal attribution metric should satisfy; we then compare the behavior of four proposed metrics in this setting and show how they are sub-optimal for the task of estimating the importance of prunable units; 2) based on these observations, we propose to estimate the contribution of each unit using Shapley values \cite{shapley1953value}, a classic result from cooperative game theory that is uniquely characterized by a collection of desirable properties; 3) we demonstrate empirically that Shapley values can identify units having a negative impact on the performance of a model and reduce the harm induced by structured pruning.

\section{Background}
\label{sec:background}

Consider a feed-forward neural network $\vec{f}$, composed of a chain of $L$ layers, each performing a (non-)linear transformation $\vec{f}^{(l)}$ on the activation $\vec{z}^{(l-1)}$ of the previous layer:
\begin{subequations}
\begin{eqnarray}
    \vec{f}(\vec{x}) = (\vec{f}^{(1)} \circ \vec{f}^{(2)} \circ ... \circ \vec{f}^{(L)}) (\vec{x}) \\
    \vec{z}^{(l)} = \vec{f}^{(l)}(\vec{z}^{(l-1)}); \quad \vec{z}^0 = \vec{x},
\end{eqnarray}
\end{subequations}
where $\vec{x}$ is an input example fed into the network. In practice, multiple inputs $\vec{x}_i \in \vec{X}$ are usually processed in parallel.

When $\vec{f}^l$ is a fully-connected layer, the activation  $\vec{z}^l \in \mathbb{R}^n$ is a vector where each element $z_1^l,...z_n^l$ represents the output of a neuron after the non-linearity (e.g. the ReLU) is applied. In this case, structured pruning is performed by removing one or more of the units $z_i^l$, which implicates removing the corresponding rows from the 2-dimensional parameter tensor $\vec{w}^l \in \mathbb{R}^{n\times m}$ as well as all the parameters interacting with $z_i^l$ on the following layer $\vec{f}^{l+1}$. 
In the case of convolutional layers, the activation  $\vec{z}^l \in \mathbb{R}^{n\times w\times h}$ is a multi-variate tensor where $n$ is the number of output channels while $w$ and $h$ represent the width and height of these channels respectively. In this case, we are interested in removing one or more channels $\vec{z}^l_i \in \mathbb{R}^{w\times h}$, which allows us to remove the corresponding filters from the parameters of the layer. 
As for dense layers, the parameter tensor of the following layers will be also shrunk accordingly. Notice that, for both fully-connected and convolutional layers, the pruning happens along a single dimension of the activation tensor. The elements $z_i^l$ along this dimension, which together constitute the activation $\vec{z}^{l}  = [z_1^l, ...z_n^l]$, will be referred to as ``prunable units'' in the remainder of the paper. 
Given this notation, pruning can be formulated as the problem of finding the optimal subset $P^l \in \{z_1^l, ...z_n^l\}$ of units that can be removed from each layer $l$ to either 1) reach a target computational cost while minimizing the loss in performance, or 2) minimize the computational cost within a certain budget of performance loss.

\subsection{Typical pruning algorithm}
In this work, we consider a typical iterative pruning strategy \cite{hu2016network, li2016pruning, luo2017thinet, molchanov2016pruning}.
This consists of three stages: 1) training a (large) network, 2) pruning one layer (usually starting from the outermost), 3) fine-tuning the network and repeating the steps (2-3) with the previous layer. Once all layers are pruned, the final network is fine-tuned once again. It has been empirically showed that iterating (2-3) by pruning only one layer at each iteration provides better results than pruning all layers at once, which might cause larger damage not recoverable by fine-tuning \cite{li2016pruning, luo2017thinet}.

In the setting that we analyze, performing step (2) as described above requires that each prunable activation $z_i^l$ is assigned an \textit{attribution value} $R_i^l$ indicating its contribution to the performance of the network, i.e. to the maximization of the accuracy or the minimization of the loss. Activations can be ranked based on their attributions and the pruning is performed retaining only those activations that rank among the top-$k$, with $k < n$. The pruning ratio $k/n$ is typically a hyper-parameter provided by the user \cite{hu2016network, li2016pruning, luo2017thinet, molchanov2016pruning}.

\subsection{Attribution metrics}
The choice of the attribution metric might affect the extent of the performance degradation after pruning as well as the effectiveness of fine-tuning for recovering the original performance.
In the following, we describe four attribution metrics proposed in the literature. 

\textbf{{$l_1$}-Norm of the weights.} The attribution is set equal to the absolute sum of the weights connected to each activation \cite{li2016pruning}. This metric does not consider the behavior of the model when provided with input data but is rather based on the argument that weights with a very low average magnitude will lead to small activations and that these, in turn, will have little influence on the network output.

\textbf{Average Percentage of Zeros (APoZ)}. The APoZ metric is computed as the fraction of times a ReLU unit is inactive with respect to the number of times the unit is evaluated when tested over a dataset $\mathcal{D}$ \cite{hu2016network}. The attribution scores would be then $R_i = 1 - APoZ(z_i)$ as activations with large APoZ are considered not to contribute much to the network performance.

\textbf{Sensitivity}. The attribution is set to be the norm of the gradient of the activation with respect to the network loss $\mathcal{L}$, averaged over a dataset $\mathcal{D}$ \cite{mittal2019studying}. 

\begin{equation}
    R_i^l = \frac{1}{|\mathcal{D}|} \sum_{\vec{x}, y \in D} \Big|\Big| \frac{\partial  {\mathcal{L}}(\vec{x}; y)}{\partial \vec{z}_i^l} \Big|\Big|_1
\end{equation}

\textbf{Taylor expansion}. Based on the first-order Taylor approximation of the loss function, \cite{molchanov2016pruning} suggested defining the attribution score as the gradient of the loss with respect to the activation multiplied with the activation itself.
When the activation $\vec{z}_i^l$ is multi-variate, with M being the length of the vectorized activation, each component is summed up before taking the absolute value:

\begin{equation}
    \label{eq:taylor}
    R_i^l = \frac{1}{|\mathcal{D}|} \sum_{\vec{x}, y \in D} \Bigg| \frac{1}{M} \frac{\partial  {\mathcal{L}}(\vec{x}; y)}{\partial \vec {z}_{i}^l}^T \vec{z}_{i}^l \Bigg|,
\end{equation}

\section{Axiomatic Comparison of Pruning Metrics}\label{sec:axiomatic}
We notice that the metrics described above are based on heuristics or first-order approximations of the model.
In this section, we endorse an axiomatic approach to discuss desirable properties that an attribution method should satisfy in order to perform optimal pruning and we show that existing metrics fail at satisfying one or more of these properties in a simple scenario, an observation that will lead us to a possible alternative.

\begin{figure}[!ht]
\centering
\includegraphics[width=0.35\textwidth]{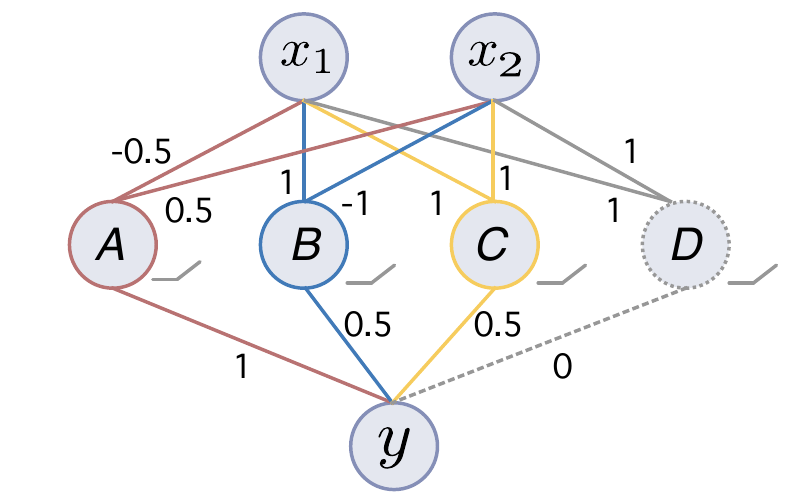}\vspace{0.15cm}
\small
\def\arraystretch{1.1}
\setlength{\tabcolsep}{3.2pt}
\noindent
\begin{tabular}[b]{c|ccccc}
\hline
  Attribution & 1-APoZ & $||w||_1$ & Sensitivity & Taylor & SV \\ \hline \hline
  A & 0.5   & \rc{1} & \rc{0} & \rc{0} & 6.25 \\ \hline
  B & 0.5   & 2     & \rc{0} & \rc{0} & 6.25\\ \hline
  C & 1     & 2     & \rc{0} & \rc{0} & 37.5\\ \hline
  D & \rc{1} & \rc{2} & 0     & 0     & 0
 \vspace{0.5cm}
\end{tabular}
\captionlistentry[table]{A table beside a figure}
\captionsetup{labelformat=andtable}
\caption{Implementation of
$y = max(x_1, x_2)$ with 4 ReLU units (
$ y = 
\color{dred} 0.5 \cdot max(0, -x_1+x_2) \color{black} + 
\color{blue} 0.5 \cdot max(0, x_1-x_2) \color{black} + 
\color{yellow} 0.5 \cdot max(0, x_1+x_2) \color{black} + 
\color{grey} 0 \color{black}$). We assume $x_1,x_2 \sim \mathcal{U}[0,10]$ and a MSE loss. Notice that the loss is zero with all inputs in the domain and (D) is the only unit that can be pruned without affecting the model. In red we highlight attributions that violate the assumptions listed in Sec.\  \ref{sec:axiomatic}. Shapley values (SV) correctly identifies (D) as having no impact on the output and sum up the expected loss gap if all units were pruned, i.e., $\mathcal{L}_{\varnothing} - \mathcal{L} = \mathcal{L}_{\varnothing} = 50$. For the analytical derivation, see the Supplementary material.}
\label{fig:example}
\end{figure}

Let us consider the model in Figure \ref{fig:example}, which illustrates a ReLU network that perfectly implements the function $y = max(x_1, x_2)$ for $x_1, x_2 >= 0$, with 4 hidden units and ReLU non-linearity. 
The configuration is such that pruning unit (D) would cause no harm to the model while removing any of the other units would cause an increment of the average loss, which we assume to be the mean-squared error between the network output and the expected output.
Notice also that, while units (A-B-C) are all useful to perform the considered task, (A) and (B) are only active for 50\% of the possible $(x_1, x_2)$ input pairs, therefore removing either of them would cause a lower expected harm than removing unit (C).
Moreover, (A-B) are symmetric: by construction, they contribute equally to the task in the given domain, as the different weights before and after the ReLU are balanced.
Given that the mechanics of this model are well-understood, we formulate four assumptions for an ideal pruning criterion:
\noindent
\begin{enumerate}
    \setlength{\itemsep}{0pt}
    \setlength{\parskip}{0pt}
    \setlength{\parsep}{0pt}
    \item unit (D) should be identified as prunable, having no impact on the output;
    \item units (A-B) should be assigned the same attribution since their perfect symmetry (and the symmetry of the input domain) suggests there should be no preference in pruning either one;
    \item units (A-B-C) should have non-zero attributions, as they are all contributing to the task;
    \item units (A-B) should be identified as having lower attribution than unit (C) because pruning either one would still allow the network to perfectly solve the task in half of the domain;
\end{enumerate}

Table \ref{fig:example} shows the attributions computed according to the metrics presented above. Notice that both the $l_1$-norm of the weights and the APoZ metrics fail at identifying the only unit (D) as prunable, thus violating (1). In this case, the problem lies in that both metrics ignore the network behavior \textit{after} the considered layer, thus producing attributions that do not necessarily reflect the real contribution of a unit to the final loss. While in our example this is due to a single weight being zero, a more complex network configuration might cause the same effect. Moreover, the weight norm also violates assumption (2) by not considering different amplification effects that might follow.

On the other hand, the Sensitivity and Taylor metrics address the aforementioned issue by considering the flow of information up to the network loss thanks to the gradient statistics.
We notice that both Sensitivity and Taylor expansion have been studied in the literature of Explainable Artificial Intelligence (XAI) as metrics to produce \textit{local explanations} of a \ac{DNN} behavior, i.e., to identify which input features to a network are mostly affecting the observed output \cite{simonyan2013deep,shrikumar2016not,ancona2019gradient}.
In fact, this connection should come at no surprise given that the problem of quantifying the influence of one input feature is not dissimilar to the one of quantifying the influence of an internal activation (with the only caveat that pruning requires an attribution score averaged over multiple possible inputs while local explanations are generated to explain a specific input-output pair).
Several works also demonstrated the limitations of pure gradient-based metrics for the problem of quantifying the importance of features, as the information provided by the gradient is only valid on a local neighborhood of the considered input \cite{sundararajan17a, shrikumar17a}. The example in Fig.\ \ref{fig:example} illustrates one possible failure case due to the gradient locality, where units that are necessary for solving the task are identified as having zero contribution as their gradients happen to be locally zero, thus violating our assumption (3).

While the one we considered is a simple toy example, assumptions (1-4) arguably constitute an important guideline towards the development of principled pruning techniques for more complex models. In the next section, we will discuss a provably unique solution in this direction.

\subsection{Shapley value attributions}
Given the activations at one specific layer $\vec{z}^{l}$, the output (and the loss) of the model does not depend on the activations of the previous layers. Therefore, we can write the network loss as a function of $\vec{z}^{l}$:
\begin{equation}
    \Tilde{\mathcal{L}}(\vec{z}^{l}; y) :=  \mathcal{L}((\vec{f}^{(l+1)} \circ  ... \circ \vec{f}^{(L)})(\vec{z}^{(l)}); y) = {\mathcal{L}}(\vec{x}; y)
\end{equation}
This formulation helps underlying the fact that the loss only depends on the units in $\vec{z}^{l}$, therefore it makes sense to quantify the marginal contribution of each of them to the \textit{loss gap} that exists between the extreme case where all units are removed (random output, large loss value) and the case where all units are left to contribute to the output.

This is a well-study problem in the literature of cooperative game theory.
Consider a set of $N$ players $P$ and a function $\vec{\hat{f}}$ that maps each subset $S \subseteq P$ of players to a real number, modeling the outcome of a game when players in $S$ participate in it. Shapley Value \cite{shapley1953value} is one way to quantify the marginal contribution of each player to the result $\vec{\hat{f}}(P)$ of the game when all players participate. For a given player $i$, its Shapley value can be computed as:
\begin{equation}
    R_i = \sum_{S \subseteq P \setminus \{i\}} \frac{|S|!(|P| - |S| - 1)!}{|P|!} [\vec{\hat{f}}(S \cup \{i\}) - \vec{\hat{f}}(S)]
\label{eq:shapley}
\end{equation}
The Shapley value for player $i$ defined above can be interpreted as the \textit{average marginal contribution} of player $i$ to all possible coalitions $S$ that can be formed without it.
Back to our problem, we can now model each prunable unit $z_1^{l}, ..., z_n^{l} \in P$ as one player and the function $\vec{\hat{f}}(S)$ as the network loss. While $\vec{\hat{f}}$ is a set function, we can replace $\vec{\hat{f}}(S)$ in Eq.\ \ref{eq:shapley} with $\Tilde{\mathcal{L}}(\vec{z}_S^{l})$, where $\vec{z}_S^{l}$ indicates the original activation vector $\vec{z}^{l}$ where all features not in $S$ have been removed.
Notably, it can be proved that Shapley value is the \textit{only}\footnote{Assuming a binary setting where activations are either enabled or disabled. When players can have different levels of demand, continuous variants of Shapley Value like Aumann-Shapley and Serial Cost methods also satisfy these properties \cite{friedman2004pathmethods}.} way of assigning attributions to players that satisfies the following four properties \cite{shapley1953value}, which we present formulated in our context:

\textbf{Null player.} \textit{If the loss does not depend on a particular activation, then its attribution is zero.} This property guarantees that our assumption (1) in Section \ref{sec:background} is satisfied, identifying unit (D) as having zero attribution.

\textbf{Symmetry.} \textit{If the loss function depends on two activations but not on their order (i.e. the values of the two activations could be swapped, never affecting the loss), then the two activations receive the same attribution.}
This property guarantees that the assumption (2) in Section \ref{sec:background} is satisfied, with units (A) and (B) receiving the same attribution.

\textbf{Linearity.} 
\textit{If the loss function $f$ can be seen as a linear combination of the loss functions of two sub-networks (i.e. $f = a \times f_1 + b\times f_2$), then any attribution should also be a linear combination, with the same weights, of the attributions computed on the sub-networks, i.e. $R_i^l(\vec{z}|f) = a\times R_i^l(\vec{z}|f_1) + b \times R_i^l(\vec{z}|f_2)$, where $R_i^l(z|f)$ denotes the attributions for the DNN that implements the loss $f$.} 
This is a natural property to expect, justified by the need of preserving linearities within the function being explained \cite{shapley1953value, sundararajan17a}.

\textbf{Efficiency.} \textit{Attributions sum up to the difference between the loss evaluated when all activations are enabled and the loss evaluated when all activations are pruned, i.e. $\sum_{i=1}^n R_i^l  = \Delta\mathcal{L} =  \Tilde{\mathcal{L}}(\vec{z}_{\varnothing}^{l}; y) - \Tilde{\mathcal{L}}(\vec{z}^{l}; y)$.}

In our context, this property is particularly interesting because it guarantees that, if $\Delta\mathcal{L} > 0$, then some units will be assigned a non-zero attribution, which is not necessarily true for gradient-based methods, as we have discussed before.
Moreover, it allows us not only to produce a ranking based on Shapley value but also to assign a well-defined meaning to the magnitude of the attributions with respect to the loss gap: $R_i^l$ is the expected loss increment that would occur when unit $z_i$ is pruned on top of a random subset of other units. This property gives us precise guarantees for the partitioning of the attributions, which is necessary for our assumptions (3) and (4).

The results in Table \ref{fig:example} shows that Shapley value attributions are the only that satisfy the assumptions (1-4) in our toy example, overcoming the limitations of other heuristics.

\subsection{Computational considerations}
It is immediately evident that computing Eq.\ \ref{eq:shapley} exactly would require $\mathcal{O}(2^{|\vec{z}^{l}|})$ network evaluations, where $|\vec{z}^{l}|$ indicates the number of prunable units on layer $l$. Intuitively, this is required to evaluate the contribution of each activation with respect to all possible subsets that can be enumerated with the other ones.
Clearly, the exact computation of Shapley values is computationally unfeasible for more than 15-20 activations.
On the other hand, several sampling-based approximation methods have been proposed \cite{castro2009, strumbelj2010efficient, datta2016algorithmic} as well as approximations specifically designed for \acp{DNN} \cite{ancona2019explaining}. In our experiments, we have chosen to use a sampling algorithm \cite{castro2009} because it is easy to implement and provides an unbiased approximation of the Shapley values in $\mathcal{O}(K|\vec{z}^{l}|)$, where $K$ is the number of samples taken. The full pruning algorithm in pseudo-code is provided in the Supplementary Material. Our PyTorch implementation is available online\footnote{https://github.com/marcoancona/TorchPruner}.
While several loss evaluations are necessary, notice that 1) no back-propagation through the network is needed, thus allowing the processing of larger batches in parallel and 2) when processing the activations of layer $l$, the forward-pass can be limited to layers $\vec{f}^{l+1}, ..., \vec{f}^L$ with a significant saving in computations. 
As an example, on VGG16 with 15M parameters trained on CIFAR-10, computing approximate Shapley values ($K=5$) for all activations of a single layer on 100 input images takes between 4 and 22 seconds on a single Titan X GPU. The actual time depends on the number of activations and the position of the layer in the model. While this is significantly more than for other attribution metrics, we argue that this time is still negligible compared to training the original model. Moreover, more sophisticated approximation methods, like DASP \cite{ancona2019explaining}, might further speed-up the computation.

\subsection{The role of sign for attributions}
\label{sec:sign}
The reader might have noticed that all attribution methods described in Section \ref{sec:background} provide positive attributions, i.e., they indicate the magnitude of the contribution of each prunable unit to the output but not the direction of this contribution. 
On the contrary, Shapley values as described by Eq. \ref{eq:shapley} can take negative values. In fact, we argue there is no reason to assume that one unit could not have a negative attribution when it degrades the overall performance of the network. In these cases, a signed attribution method would be beneficial to make sure harmful units are pruned first.

In order to verify that Shapley values can indeed identify negatively-contributing units, we considered a randomly initialized fully-connected \acp{DNN} with two hidden layers of 2048 units each and equipped with a LeakyReLU non-linearity. 
We computed the Shapley values for all prunable units using the MNIST \cite{lecun1998mnist} and CIFAR-10 \cite{krizhevsky2009learning} datasets on this network targeting the average loss of 10,000 samples taken from the respective training sets. Without performing any training, we then pruned all units with negative average attribution, leaving the remaining parameters at their initialization value. We run this process with five random initializations and the best pruned networks increased its \textit{test} accuracy to $50.01\%$ on MNIST and to $23.25\%$ on CIFAR-10, a significant improvement over the initial baseline ($\approx 10\%$). This is remarkable considering that the weights have not been trained at all. In this case, the pruning acts as optimization for the given task. The phenomenon suggests that 1) given a large number of parameters, even a random initialization might produce sub-graphs that partially solve the given task with random weights, and 2) there might exist units that play a negative role for the given task, such that pruning them would increase the overall performance. Given these observations, we argue that an optimal pruning strategy should rely on signed attributions methods.

In practice, however, using the absolute value of the attributions was found producing better results \cite{molchanov2016pruning, lee2018snip}.
Examining the Shapley values of individual units across several inputs, we found that the same units can have a large positive attribution on some of the examples while showing a large negative attribution on others (details in the Supplementary material).
In these cases, a naive \textit{average} aggregation over the samples might assign a low, or even negative, attribution to some units without general consensus. 
The resulting ranking promotes the pruning of units that might be particularly important to handle a significant part of the input distribution. 
We argue this is not desirable, as it is important to preserve the network performance on most of the input data. This could be achieved using a more conservative aggregation function.
We found that a simple ranking based on mean \textit{plus twice the standard deviation} of the attributions is more robust and produces significantly better results without sacrificing the ability to detect a negative impact on the output when there is a large consensus among the input samples.
We use this ranking for all experiments that follow.

\section{Empirical evaluation}
\label{sec:results}
Having discussed the theoretical properties of Shapley values, we now present two experiments that will aid the comparison with the existing metrics presented in Section \ref{sec:background}.  Additional results and details on the experimental setup can be found in the Supplementary material.

\subsection{Layer-wise pruning robustness}

\begin{figure*}[ht]
\centering
\includegraphics[width=1.0\textwidth]{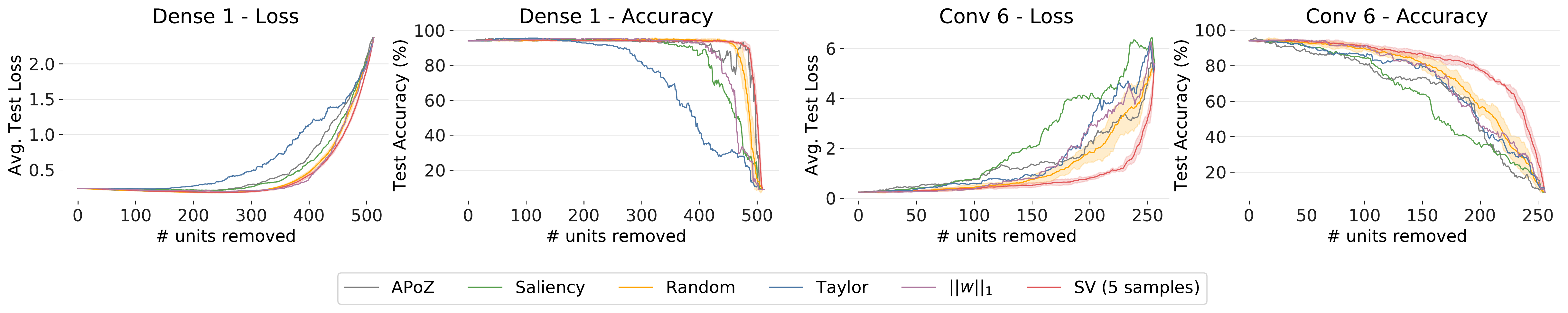}
\caption{Test loss and accuracy on VGG16 and CIFAR-10, as the units of two of its layers are sequentially pruned. For each curve, the pruning order follows the ranking computed with the corresponding attribution metric. For Shapley values (SV) the ranking is computed using the conservative aggregation as described in Sec.\ \ref{sec:sign}. As Random and SV are not deterministic, we indicate the standard deviation over 5 trials with a shadowed area. Better attribution methods are expected to induce a slower deterioration of the performance.}
\label{fig:auc}
\end{figure*}

In order to compare the different metrics and factor-out other variables that might affect the result, we test the robustness of a model by pruning each of its layers independently and measuring how the performance decreases.
This analysis is similar in spirit to the ``initial damage'' evaluation by \cite{mittal2019studying}, except that we prune one unit at a time and measure its direct effect on the network output, layer by layer. 
Fig. \ref{fig:auc} illustrates the test loss and accuracy while convolutional filters are sequentially removed at one specific layer of a VGG16 model \cite{simonyan2014very} trained on CIFAR-10 with $93.3\%$ initial test accuracy. Filters are pruned following the ranking given by different attribution methods. If an attribution method better identifies the least important filters, we expect the loss (accuracy) to increase (decrease) slower while these units are removed. To obtain a quantitative metric across all layers, we compute the Area Under the Curve (AUC) of the loss as follows:
%
\begin{equation}
    AUC(\vec{X}, \vec{y}) = \frac{1}{N} \sum_{l=1}^{L} \sum_{i \in \mathcal{O}^l}  {\mathcal{L}}^i(\vec{X}; \vec{y}) - {\mathcal{L}}(\vec{X}; \vec{y}), 
\end{equation}
where $N$ is the number of total prunable units in the network, $L$ is the number of prunable layers and the inner sum is over the indices $i$ of the prunable units, sorted at each layer according to the ranking $\mathcal{O}^l$ provided by the attribution metric. ${\mathcal{L}}^i$ indicates the loss when all units up to $i$ have been pruned. 
The AUC aggregated results in Table \ref{table:auc} show that pruning based on Shapley value produces less performance degradation than other metrics on average.

\begin{table}[!ht]
\centering
\small
\def\arraystretch{1.1}
\setlength{\tabcolsep}{3.2pt}
\noindent
\begin{tabulary}{\linewidth}{c|cccccc}
\hline
  AUC & APoZ & $||w||_1$ & Sensitivity & Taylor & Random & Shapley values \\ \hline \hline
  FMNIST  & 0.30   & 0.19  & 0.24 & 0.20 & 0.29$\pm$0.02 & \textbf{0.11$\pm$0.00} \\\hline
  CIFAR-10 & 0.56   & 0.47  & 0.47 & 0.47 & 0.48$\pm$0.01 & \textbf{0.31$\pm$0.01} 
\end{tabulary}
\caption{Test loss AUC, summed over all prunable layers and normalized by the total number of units. A smaller value indicates that the performance degrades slower. Standard deviation over 5 runs is reported when applicable.}
\label{table:auc}
\end{table}

\subsection{Pruning in low-data regime}
A recent work showed that fine-tuning a network after pruning leads to the same recovery in performance regardless of the pruning criterion that had been used \cite{mittal2019studying}. The authors found that, given enough training data and fine-tuning time, the network plasticity allows closing the performance gap between different metrics, including random pruning. We observed the same phenomenon in our experiments, as reported in the Supplementary material: as the amount of data available for fine-tuning increases, the performance gap between different metrics becomes smaller.

Here we focus instead on a low-data regime, when either the original training data is not available or when transfer learning is used to adapt a pre-trained network to a smaller dataset. 
We consider two VGG16 models, the first pre-trained on CIFAR-10 (15M parameters) and the second pre-trained on ImageNet and fine-tuned on the more challenging Caltech-UCSD Birds 200 (CUB-200) dataset \cite{welinderEtal2010cub200} (135M parameters). We randomly take aside 100 examples from each test set, since these have not been seen by the network during training, and we use these samples to compute the attribution metrics. We keep the remaining examples in the original test set for the final performance evaluation. 
Then we prune the network sequentially, one layer at the time, with no fine-tuning as in \cite{yeom2019pruning}. As recommended by others \cite{mittal2019studying}, we do not prune the first four convolutional layers, which have very few filters, because this leads to negligible computational saving with a large drop of accuracy. %
After pruning each layer, we measure the network accuracy on the large test set we kept aside.

\begin{figure*}[h]
\centering
\includegraphics[width=1.0\textwidth]{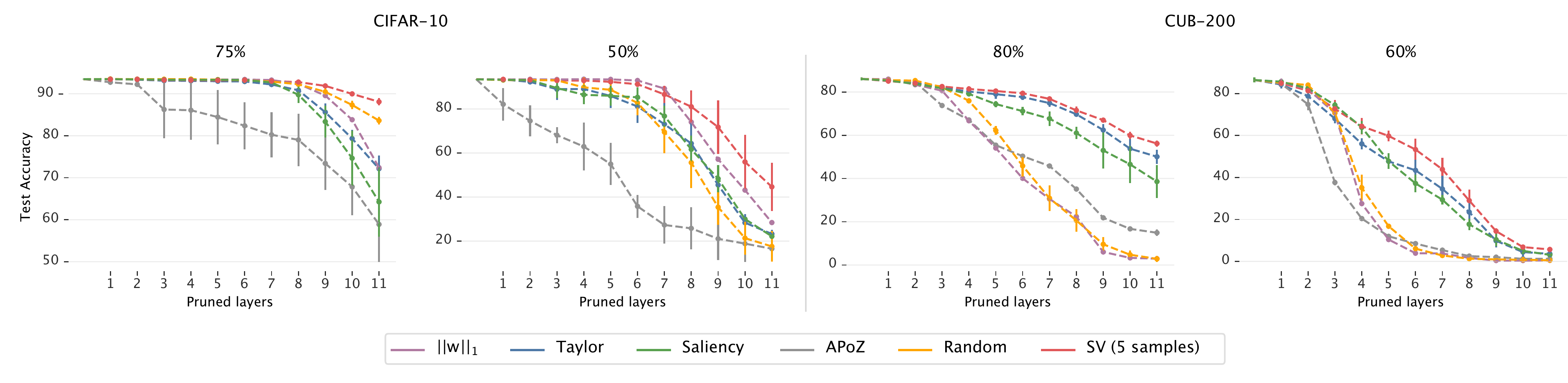}
\caption{
Test accuracy (and standard deviation over 3 runs) for VGG16 trained on CIFAR-10 and CUB-200, after pruning in low-data regime.
We prune 25\% and 50\% of units on CIFAR-10, and 20\% and 40\% of units on the more challenging CUB-200.
Notice how the second best method varies depending on the dataset and the pruning ratio, while Shapley values consistently outperforms the other metrics.
Vertical scale adjusted for readability. Best seen in electronic form.
}
\label{fig:low_data_notune}
\end{figure*}

The results are reported in Figure \ref{fig:low_data_notune} for two levels of pruning. First, we notice that at low pruning ratio the network performance is well-retained by pruning up to layer 7, after which the accuracy decreases fast as we prune the last four layers. 
Notably, the ranking of the different attribution metrics varies largely among datasets and pruning ratios. For example, while random pruning is particularly effective when removing a small amount of units on CIFAR-10, second only to Shapley values, it performs poorly on CUB-200.
Conversely, Taylor is the second best metric on CUB-200, in agreement with the results of \cite{molchanov2016pruning}, but it only performs average on the CIFAR-10 dataset. Shapley values shows consistently best results across the four experiments.

\section{Other Related Work}
To increase the pruning effectiveness, sparsity can be artificially promoted during training with suitable regularization, either in unstructured \cite{srinivas2017training, louizos2018learning} or structured \cite{lebedev2016fast, liu2017learning} settings. While we do not enforce any prior, these could be used with our pruning pipeline encouraging units with low absolute attribution.
Other works proposed to find sparse convolutional networks by greedily optimizing for the reconstruction error on the following layer \cite{luo2017thinet} or as a solution of a LASSO regression \cite{he2017channel}. A different approach is to tackle sparsity with meta-learning  \cite{chin2018layer} and reinforcement-learning \cite{he2018amc}. In contrast, we analyze networks trained in a traditional fashion and focus our attention on the problem of estimating the importance of internal units for the solution of a given task.

A recent work draws the connection between pruning and explainability \cite{yeom2019pruning}. The authors propose Layer-wise Relevance Propagation (LRP) as an attribution metric for pruning. While this paper focused on attribution methods that do not require a dedicated layer implementation, we provide an axiomatic comparison with LRP in the Supplementary material.

Recently, \cite{frankle2018the} showed that most \acp{DNN} might contain smaller sub-networks (\textit{winning tickets}) that, thanks to their fortuitous weights initialization, can achieve the same performance of a full model when trained in isolation. 
These results have been improved by \cite{lee2018snip,wang2020picking}, while \cite{zhou2019deconstructing} showed that pruning on a random-initialized network can improve its performance well-beyond chance with no training. 
All these works focused on unstructured pruning. In contrast, we show that in a structured pruning setting it is also possible to find harmful units thanks to signed attribution methods.

Besides pruning, distillation \cite{ba2014distillation, hinton2015distilling}, quantization \cite{gong2014compressing, chen2015compressing, han2015deep} and low-rank approximation \cite{denton2014exploiting} are other possible strategies for reducing the network inference cost. Architectures that are specifically engineered to be one order of magnitude smaller than traditional ones are also possible \cite{iandola2016squeezenet, howard2017mobilenets}. While we focus on pruning, these strategies can be often combined for better results \cite{han2015deep}.

\section{Conclusions}
In this work, we first compared four popular heuristics to estimate the contribution of the internal units of a \ac{DNN} with respect to its performance. We discussed the theoretical limitations of these metrics, highlighting failure cases on a simple, interpretable model. 
Then we proposed the use of Shapley values, a classic solution from cooperative game theory, to provably overcome such limitations. 
Finally, we discussed the role of the sign for attribution methods and showed empirically that structured pruning based on Shapley values causes less initial harm and better retains the network performance.
As research on this area continues, we hope that this work will encourage the exploration of principled pruning metrics as well as faster approximation techniques for Shapley values tailored to this application.




\begin{ack}
The authors would like to thank Dr. Tobias G{\"u}nther for the support and useful discussion.
\end{ack}


\bibliography{paper}
\bibliographystyle{dinat}

\end{document}


\date{}
\maketitle

\appendix

\section{The role of sign for attributions}

\begin{figure}[!ht]
\centering
\includegraphics[width=0.7\linewidth]{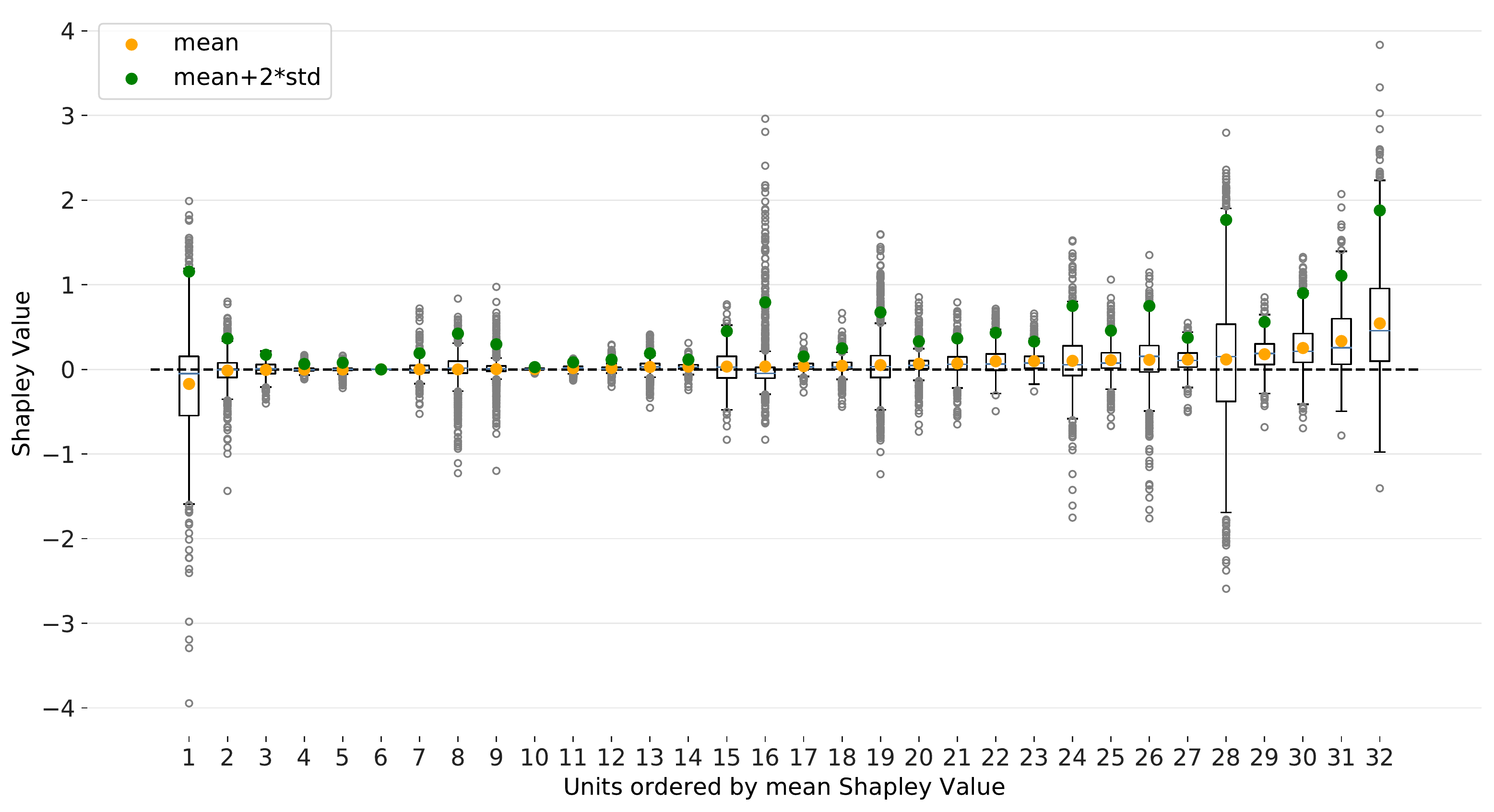}
\caption{Shapley value distribution, over 1000 inputs, for the 32 prunable filters of the first convolutional layer of a CNN trained on the Fashion-MNIST dataset \cite{xiao2017fmnist}. The units are sorted by their \textit{average} Shapley value (yellow marker). Even if the attribution is negative on average, some units might have a positive impact on a significant portion of the input data. A ranking based on both mean and standard deviation (green marker) is more conservative in pruning units with high variance.}
\label{fig:sv_ditribution}
\end{figure}

Previous works found that using the absolute value of the attributions produces better results \cite{molchanov2016pruning, lee2018snip}.
In particular, \cite{molchanov2016pruning} showed that a signed version of the Taylor expansion performs worse, attributing this surprising result to the instability that accumulates due to the large absolute changes induced by pruning units with large negative attribution. 
We argue that the apparent superior performance of unsigned attributions might also have a more pragmatic explanation. 
Examining the Shapley values of individual units across several inputs, we found that the same units can have a large positive attribution on some of the examples while showing a large negative attribution on others.
Figure \ref{fig:sv_ditribution} illustrates this phenomenon by showing the distribution of the Shapley values for each prunable unit of a trained convolutional layer among 1000 training samples. Notice how the unit with lowest (and negative) average attribution has, in fact, a large variance. 
In these cases, a naive \textit{average} aggregation over the samples might assign a low, or even negative, attribution to some units without general consensus. 

\section{Pruning procedure}
In this section we describe our pruning procedure in detail and discuss some technical differences between \textit{masking} activations (i.e., simulated pruning) and \textit{slicing} the network parameters (i.e., actual pruning).

Consider a feed-forward neural network $\vec{f}$, composed of a chain of $L$ layers, each performing a (non-)linear transformation $\vec{f}^{(l)}$ on the activation $\vec{z}^{(l-1)}$ of the previous layer:

\begin{subequations}
\begin{eqnarray}
    \vec{f}(\vec{x}) = (\vec{f}^{(1)} \circ \vec{f}^{(2)} \circ ... \circ \vec{f}^{(L)}) (\vec{x}) \\
    \vec{z}^{(l)} = \vec{f}^{(l)}(\vec{z}^{(l-1)}); \quad \vec{z}^0 = \vec{x},
\end{eqnarray}
\end{subequations}
where $\vec{x}$ is a input example fed into the network.
Before applying a non-linearity, Linear and Convolutional layers can be seen as an affine transformation of the previous layer activations:

\begin{equation}
  z_j^{(l)} = \sum_i w_{j1}^{(l)} z_i^{(l-1)} + b_j^{(l)}.  
\end{equation}

Actual pruning on layer $l$ requires to slice both $\vec{w}^{(l)}$ (along the first dimension) and $\vec{b}^{(l)}$ on the same indices. This will produce an activation vector $\vec{z}^{(l)}$ with fewer elements than the original. Alternatively, it is possible to \textit{mask} the elements that would otherwise be removed without affecting the number of elements in the activation vector. Notice that masking does not reduce the computational cost of the network but it is usually more easily implemented because all subsequent layers would accept the new input without the need to prune their parameters accordingly.

While slicing and masking are equivalent if another affine transformation follows in the computational graph, the following cases need to handled with care:
\begin{itemize}
    \item \textbf{Non-linearity}. Non-linear activations that map zero to a value different than zero (e.g. Sigmoid, Softplus) would produce different results for slicing and masking. Using masking, pruned activations are restored with a non-zero value after the non-linearity.
    
     \item \textbf{Batch Normalization}. Batch Normalization can add a non-zero bias to masked activations, thus making the result of slicing and masking differ from each other.
\end{itemize}

In order to avoid inconsistencies, for each Linear or Convolutional layer, we compute attributions and perform masking after any Batch Normalization and/or non-linear activation, if present. If a Dropout layer follows the pruned layer before the next affine transformation, we also adjust the dropout rate $p$ as $p_{new} = p_{old} * pr$, where $pr$ is the ratio between the number of units after and before pruning.

When we perform actual pruning, we also slice all the necessary parameters of the network to keep the computational graph consistent. These include the weight of the following affine transformation, weight, bias, running mean and running variance of Batch Normalization layers and the momentum tensor if used by the optimizer.

Algorithm \ref{alg:sv} shows the pruning procedure with Shapley value attributions in pseudo-code.

\begin{algorithm}[h]
   \caption{Compute SV and pruning ranking on one layer}
   \label{alg:sv}
\begin{algorithmic}[1]
   \STATE {\bfseries Input:} layer index $l$, number of Shapley value samples $K$, dataset $\mathcal{D} = (\vec{X}, \vec{y})$
    \STATE {\bfseries Output:} Shapley value attributions $\vec{R}_\mu$, unbiased ranking $\vec{Q}$ and conservative ranking $\vec{Q}_{robust}$ )
   
   \STATE $M = len(\vec{X)}$ // number of samples
   \STATE $N = |\vec{z}^l|$ // number of prunable units
   \STATE $\vec{R} = \vec{0}^{M\times N}$
   
    \STATE $\vec{Z}^l = (\vec{f}^{(0)} \circ  ... \circ \vec{f}^{(l)})(\vec{X});$
   \STATE $\vec{loss} = \Tilde{\mathcal{L}}(\vec{Z}^{l}; \vec{y});$
   
   \FOR{$j=1, ..., K$}
   \STATE $\Bar{\vec{Z}}^l = \vec{Z}^l;$
   \STATE $\vec{prevLoss} = \vec{loss};$
   \FOR{$i$ in $random\_permutations(N)$}
   
    \STATE $\Bar{\vec{Z}}^l[i] = \vec{0}$
    \STATE $\vec{newLoss} = \Tilde{\mathcal{L}}(\Bar{\vec{Z}}^{l}; \vec{y});$
    
    \STATE $\vec{R}[:, i] = \vec{R}[:, i] + (\vec{newLoss} - \vec{prevLoss})$
    \STATE $\vec{prevLoss} = \vec{newLoss}$
   
   \ENDFOR
   \ENDFOR

   \STATE $\vec{R} = \vec{R} / K$   \quad // Average over K samples

   \STATE $\vec{R}_\mu = mean(\vec{R}, axis=0);$ \quad //Mean over inputs
   \STATE $\vec{R}_\sigma = std(\vec{R}, axis=0);$ \quad // Std over the inputs
   \STATE $\vec{Q} = argsort(\vec{R}_\mu);$
   \STATE $\vec{Q}_{robust} = argsort(\vec{R}_\mu + 2\vec{R}_\sigma);$
 
\end{algorithmic}
\end{algorithm}

\FloatBarrier
\section{Derivations for the $max$ network}
Our toy network implements the function $y = max(x_1, x_2)$. We assume a mean squared error loss $\mathcal{L}$ and two independent input variables following a uniform distribution, i.e., $x_1,x_2 \sim \mathcal{U}[0,10]$. 
Since the network perfectly implements the $max$ function, the loss $\mathcal{L}$ is zero if none of the units (A-C) is pruned. Conversely, it is easy to compute the loss when all units are pruned, as the output of the network in this case is always zero:

\begin{align*}
    \mathcal{L}_{\varnothing} &= \mathbb{E}_{x,y} \big[ max(x_1, x_2) - f(x_1, x_2) \big] ^2 \\
    &= \mathbb{E}_{x,y} \big[ max(x_1, x_2) - 0 \big] ^2  \\
    &= \int_0^{10} \int_0^{10} max(x_1, x_2)^2 p(x_1)p(x_2) dx_1dx_2 \\
    &= \frac{1}{100} \bigg[ \int_0^{10} \int_0^{y}  x_2^2 dx_1dx_2 + \int_0^{10} \int_y^{10}  x_1^2 dx_1dx_2  \bigg] \ \\
    &= 50
\end{align*}

In this small example, Shapley values can be derived analytically applying the definition, i.e., enumerating all subsets of features that can be composed. As an example, the Shapley value of unit (A) can be computed as follows:

\begin{align*}
R_A &= \frac{1}{4} \bigg[ (\mathcal{L_{\{B,C\}}} - \mathcal{L_{\{A,B,C\}}}) + (\mathcal{L_{\{B\}}} - \mathcal{L_{\{A,B\}}}) \\ 
&+ (\mathcal{L_{\{C\}}} - \mathcal{L_{\{A,C\}}}) + (\mathcal{L_{\varnothing}} - \mathcal{L_{\{A\}}}) \bigg] = 6.25
\end{align*}

In this derivation, we have ignored the coalitions that include unit (D) as this has no impact on the output and does not affect the Shapley value. For the other units, the Shapley values can be derived either analytically or by exploiting the properties of Shapley values:

\begin{align*}
R_B &= R_A = 6.25, \quad \text{(symmetry)}\\
R_D &= 0, \quad \text{(null player)}\\
R_C &= (\mathcal{L}_{\varnothing} - \mathcal{L}) - R_A - R_B - R_D = 37.5 \quad  \text{(efficiency).}
\end{align*}

\section{Axiomatic comparison with LRP}
A recent work proposed Layer-wise Relevance Propagation (LRP) as attribution metric to assess the importance of the hidden units and thus guide the pruning procedure \cite{yeom2019pruning}. In this section, we compare LRP to Shapley values axiomatically.

LRP, originally developed to explain the importance of input features to the output of a neural network \cite{bach2015pixel}, produces attributions by back-propagating a quantity called ``relevance'' from one output neuron throughout the network layers up to the input. Several heuristics for the propagation rule have been proposed within the LRP framework, with empirical results often showing superior performance in identifying important features compared to first-order gradient methods such as Taylor expansion \cite{bach2015pixel, montavon2019layer}.
The algorithm by Yeom \textit{et al.} assumes ReLU non-linearities and positive pre-softmax activations. It relies then on the LRP-$\alpha_1\beta_0$-rule to propagate the attributions $R$ recursively, from one layer $l+1$ to the preceding one as follows:

\begin{equation}\label{eq:lrp}
    R_i^l = \sum_j \frac{z_i^l w_{ij}^{+}}{\sum_i z_i^l w_{ij}^{+}} R_j^{l+1},
\end{equation}

where $w^{+} = max(0, w)$. 
In the following, we discuss the different assumptions and properties of Shapley values and LRP.

\paragraph{Sign.}
As the pathways with negative weights are discarded during the back-propagation, attributions produced by LRP are always non-negative. On the contrary, Shapley values are not biased towards either positive or negative evidence.
The bias towards positive attributions is illustrated in the following example, where we consider a toy network similar to the one discussed in the main paper, this time with unit (D) influencing the output through a small negative weight. Notice that unit (D) harms the prediction of the network. While the Shapley value for unit (D) is negative, as we would expect for a unit that negatively contributes to the task, it is assigned a zero attribution by LRP.

\begin{figure}[!ht]
\centering
\includegraphics[width=0.35\textwidth]{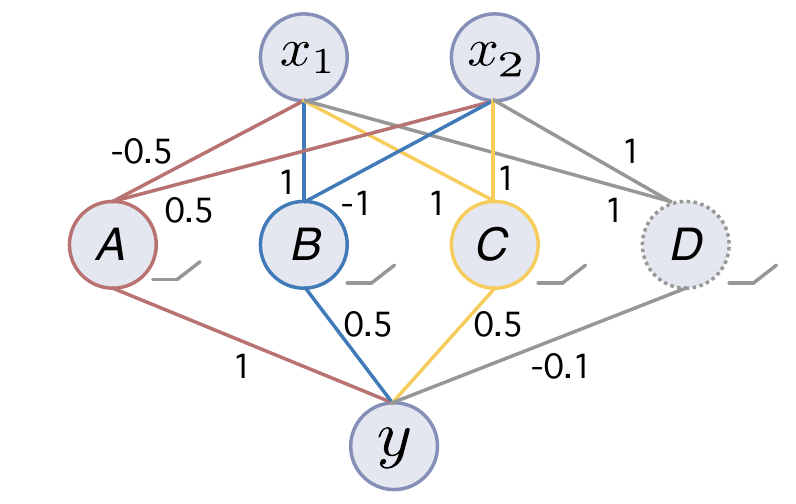}
\small
\def\arraystretch{1.1}
\setlength{\tabcolsep}{3.2pt}
\begin{tabular}[b]{c|ccccc}
\hline
  Attribution  & SV & LRP  \\ \hline \hline
  A & 7.1 & 0.7 \\ \hline
  B & 7.1 & 0.7 \\ \hline
  C & 43.4 & 4.2 \\ \hline
  D & -8.7    & 0
  \vspace{0.5cm}
\end{tabular}
\captionlistentry[table]{Comparison of SV and LRP on $max$ network}
\captionsetup{labelformat=andtable}
\caption{Implementation of
$y = max(x_1, x_2) +\epsilon$ with 4 ReLU units, where $\epsilon = -0.1 * (x_1+x_2)$ can be seen as an error caused by unit (D). The pruning of unit (D) would decrease the loss because this unit is solely responsible for the error term. While Shapley values detects the negative attribution of (D), LRP assigns zero attribution to it, as negative paths are ignored. We assume $x_1,x_2 \sim \mathcal{U}[0,10]$ and a MSE loss. Attributions computed empirically on 10,000 samples.}
\label{fig:example}
\end{figure}

\paragraph{Performance & Implementation} Compared to Shapley values, LRP is significantly faster to compute, requiring a single backward pass through the network. On the other hand, LRP requires special layers to be implemented to support the custom propagation rule, making pruning more technically demanding compared to all the methods discussed in the paper. Moreover, it assumes ReLU non-linearities and non-negative output activations.

\paragraph{Properties} With an axiomatic comparison, it is easy to see that LRP satisfies Symmmetry and Efficiency\footnote{LRP attributions sum up to the value of the target output. This is equivalent to Efficiency assuming a zero target output when all inputs are zero.} but fails to satisfy Null player\footnote{Consider the network $y = ReLU(x) - ReLU(x) + 1$. While the output $y$ does not depend on the value of $x$, LRP assigns attribution $R = 1$ to the input $x$ while it is clear that the output only depends on the bias term.} and Linearity\footnote{The property is trivially violated for any linear combination that involves negative weights.}.

\section{Layer-wise pruning robustness - Additional results}
We report the test and accuracy curves for the layer-wise robustness study on all layers of VGG16 on CIFAR-10. In these plots, we also include a comparison with the performance of Shapley values aggregated over the mean of the input examples, as well as with signed Taylor expansion, i.e. the first-order Taylor expansion metric computed according to Equation 3 but without taking the absolute value before aggregating over the input data. Both these methods underperform. We discuss a possible reason for this in Appendix A.

\begin{figure}[ht]
\centering
\begin{subfigure}
    \centering
    \includegraphics[width=0.45\textwidth]{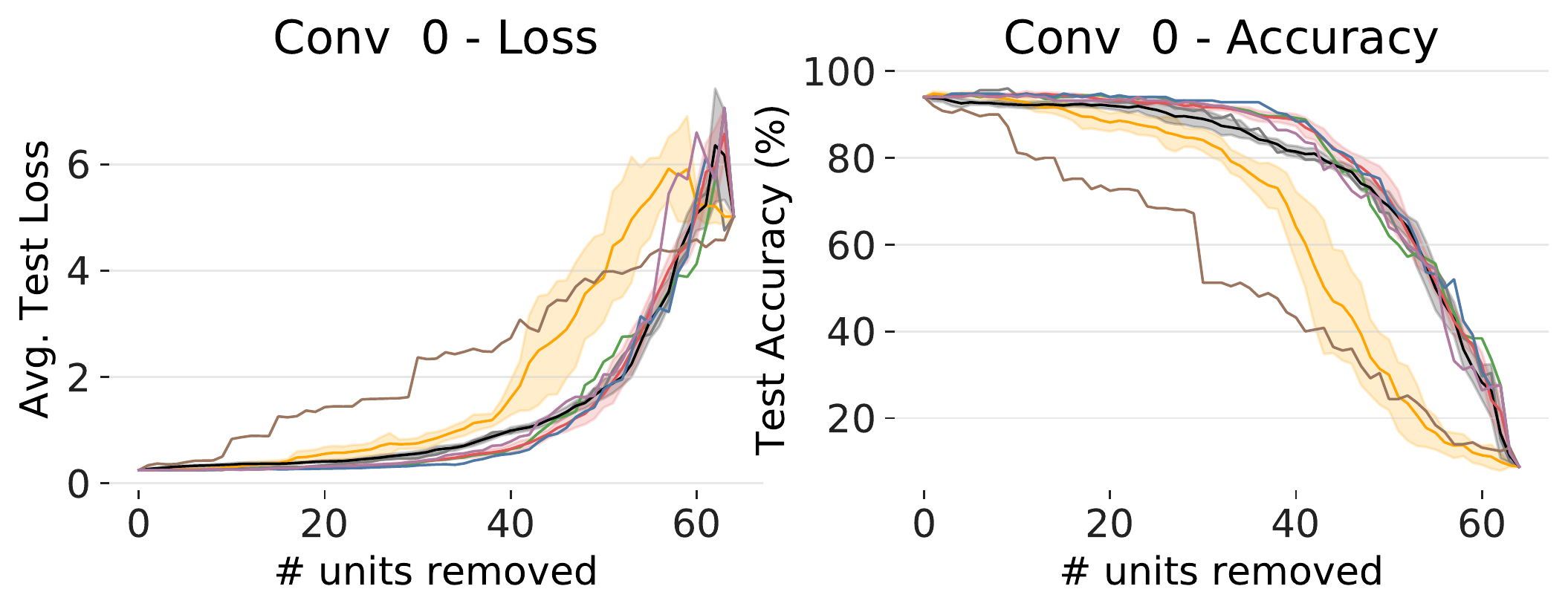}
\end{subfigure}
\begin{subfigure}
    \centering
    \includegraphics[width=0.45\textwidth]{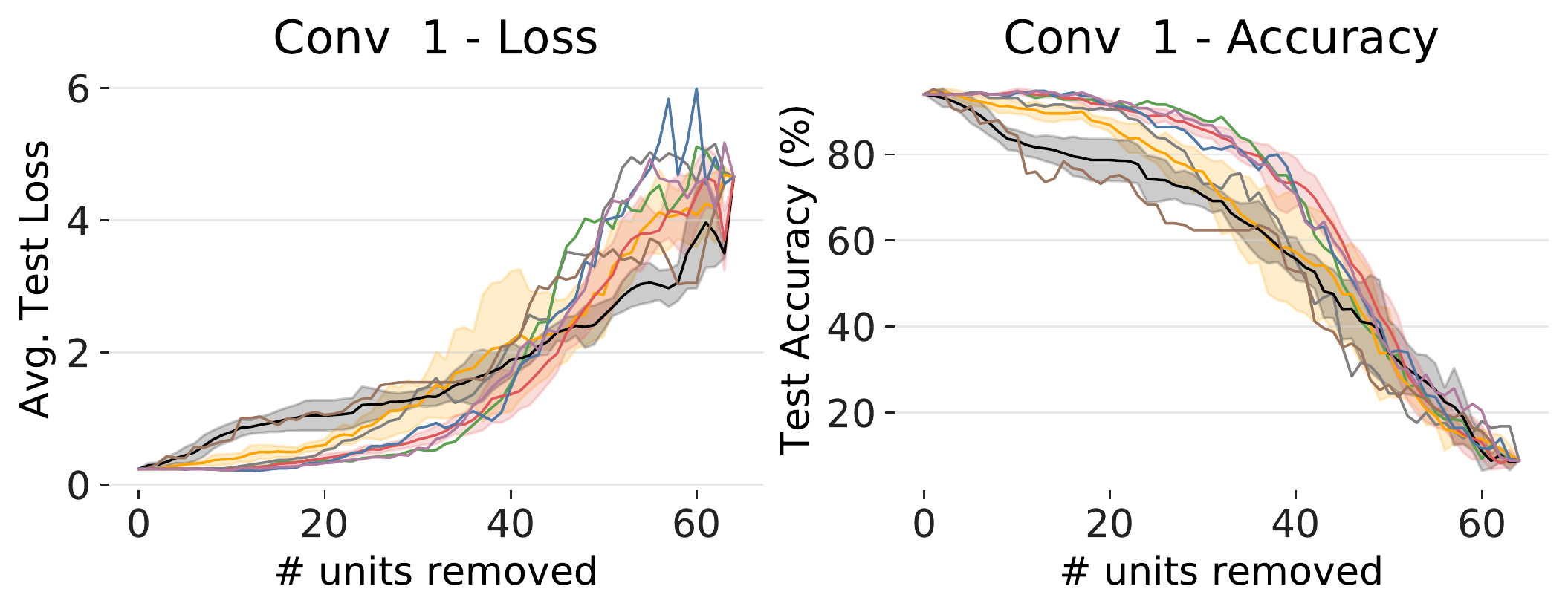}
\end{subfigure}
\begin{subfigure}
    \centering
    \includegraphics[width=0.45\textwidth]{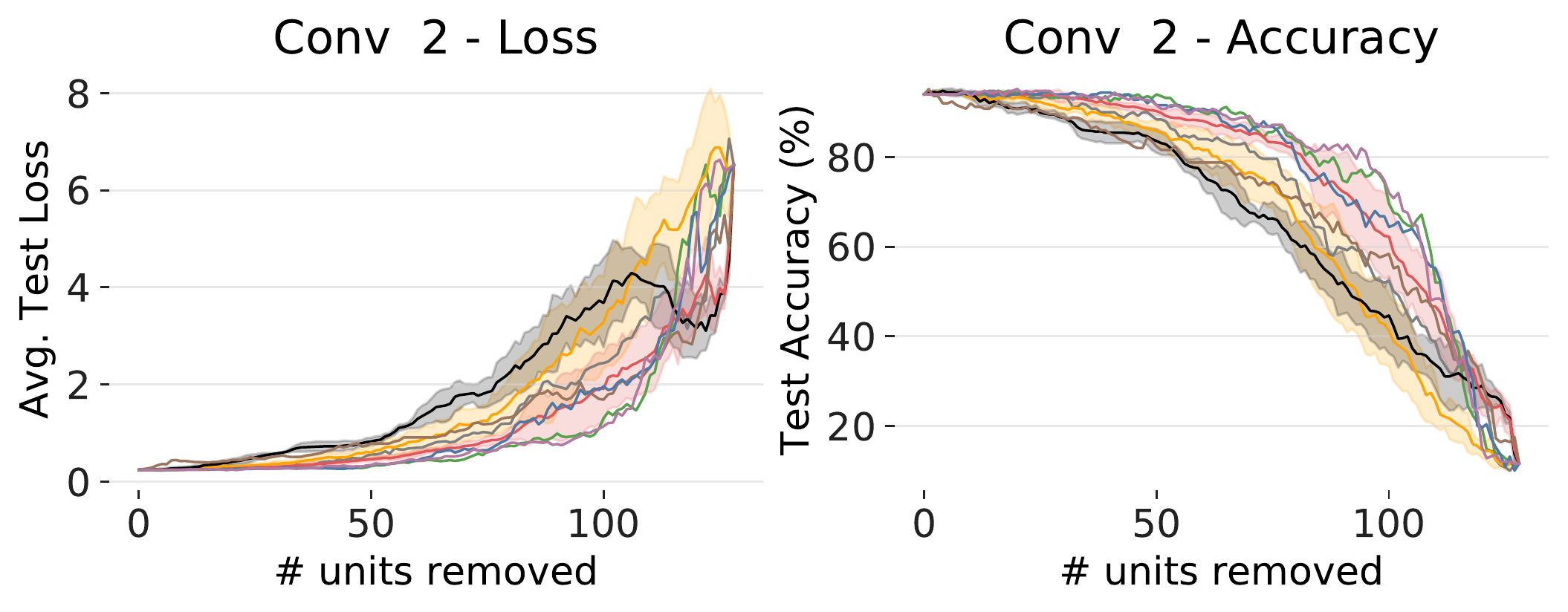}
\end{subfigure}
\begin{subfigure}
    \centering
    \includegraphics[width=0.45\textwidth]{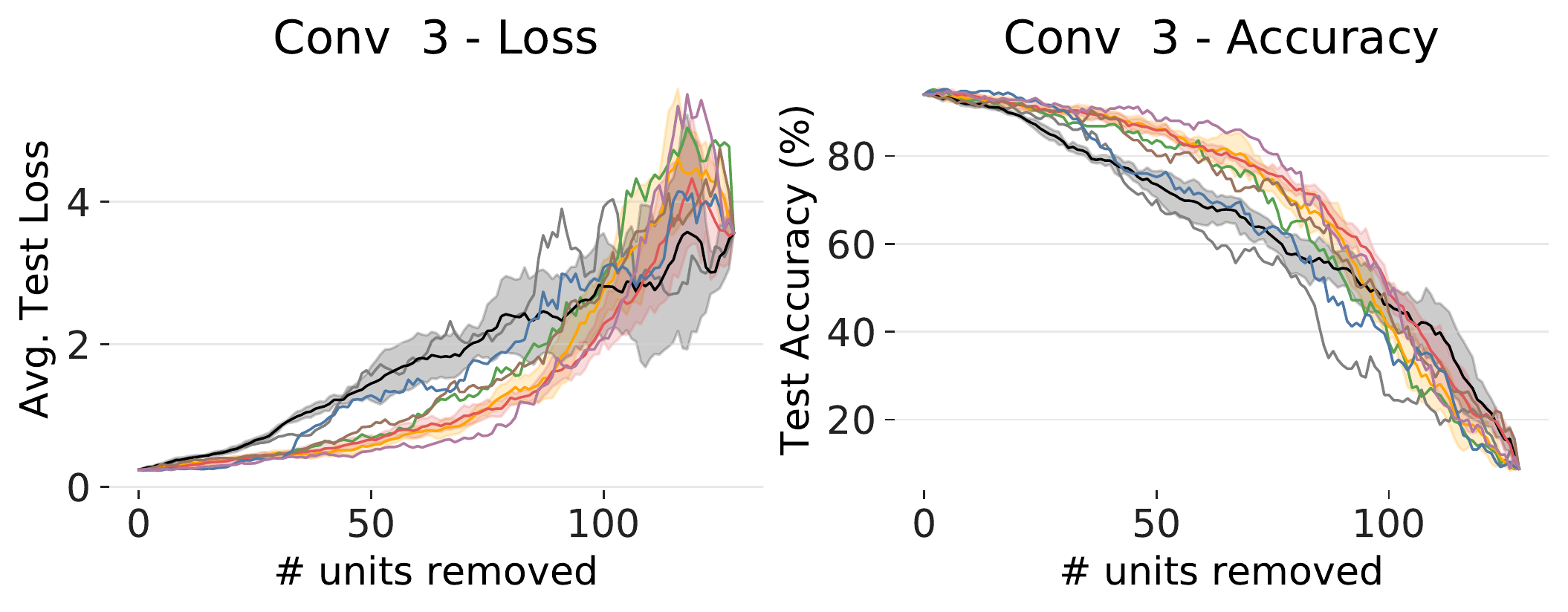}
\end{subfigure}
\begin{subfigure}
    \centering
    \includegraphics[width=0.45\textwidth]{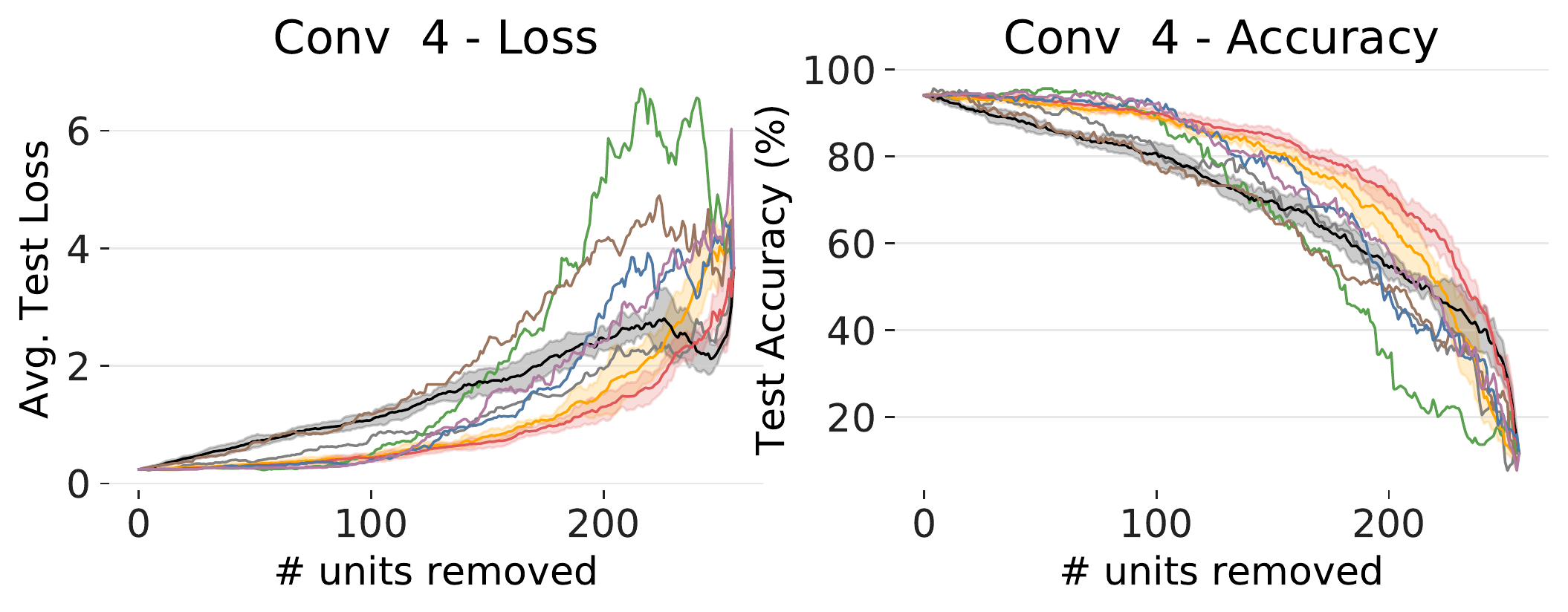}
\end{subfigure}%
\begin{subfigure}
    \centering
    \includegraphics[width=0.45\textwidth]{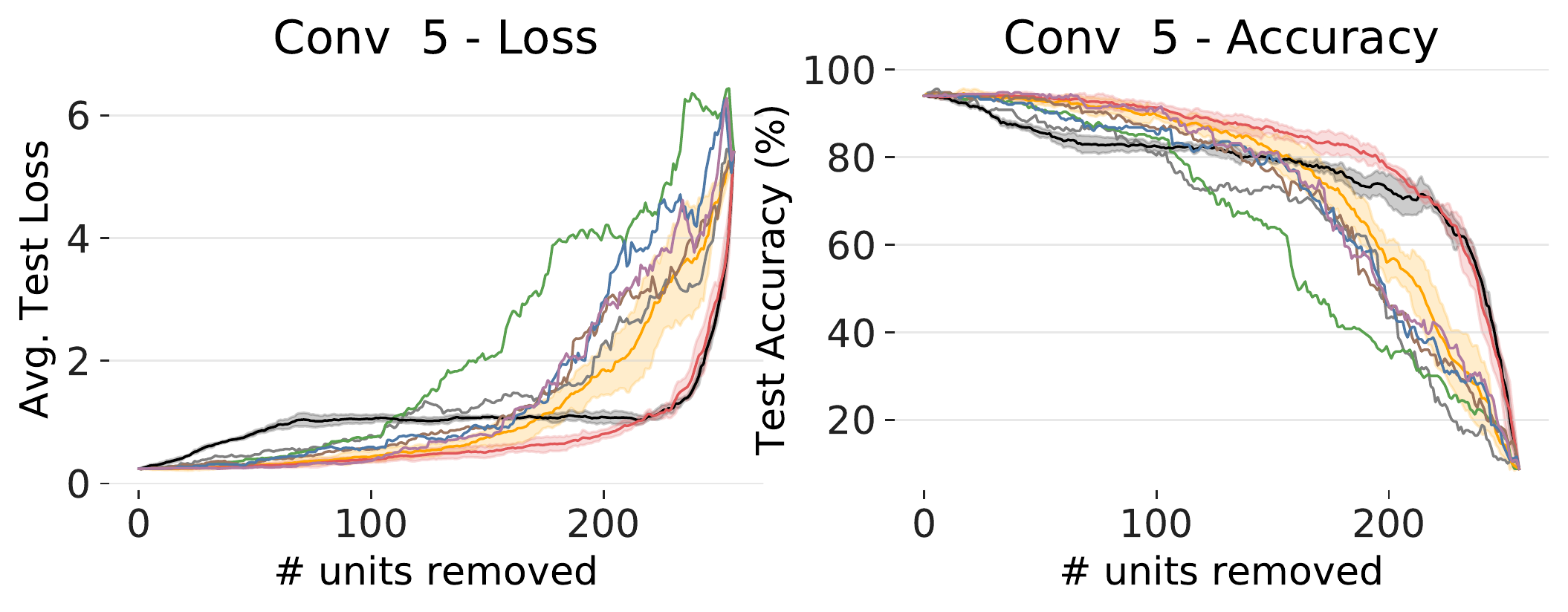}
\end{subfigure}

\end{figure}

\begin{figure}
\begin{subfigure}
    \centering
    \includegraphics[width=0.45\textwidth]{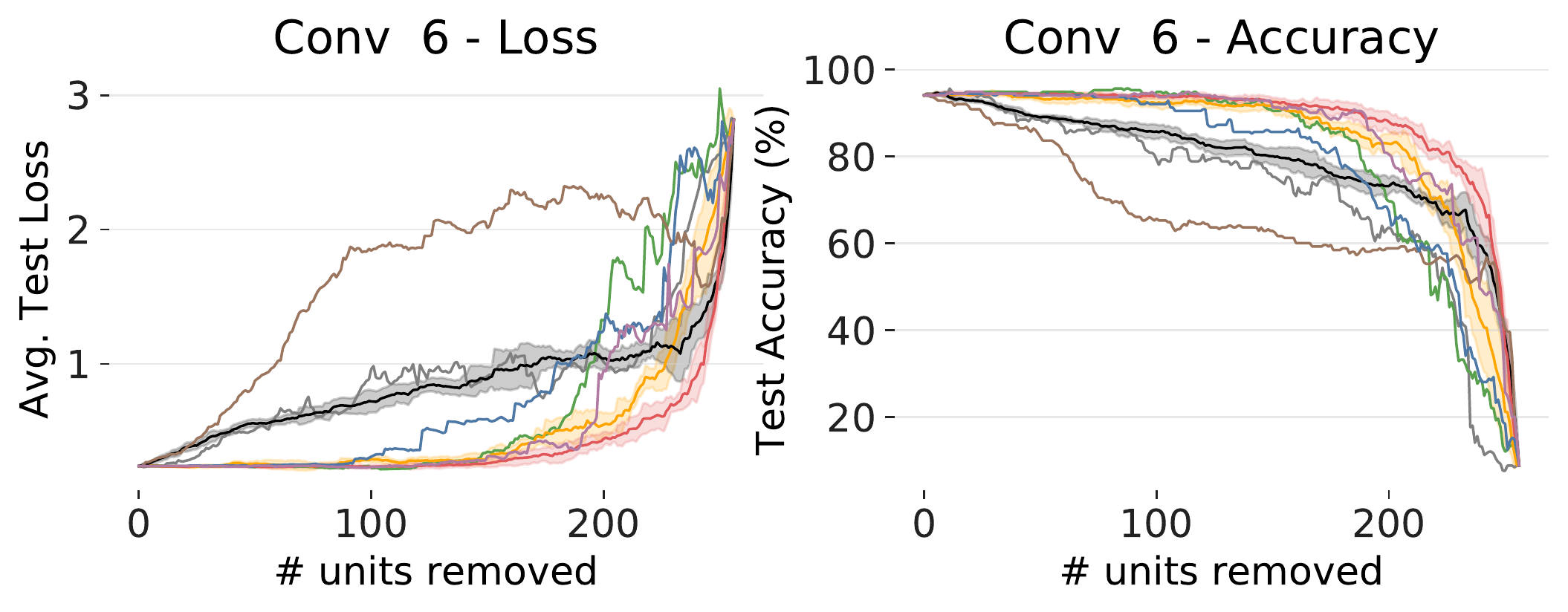}
\end{subfigure}
\begin{subfigure}
    \centering
    \includegraphics[width=0.45\textwidth]{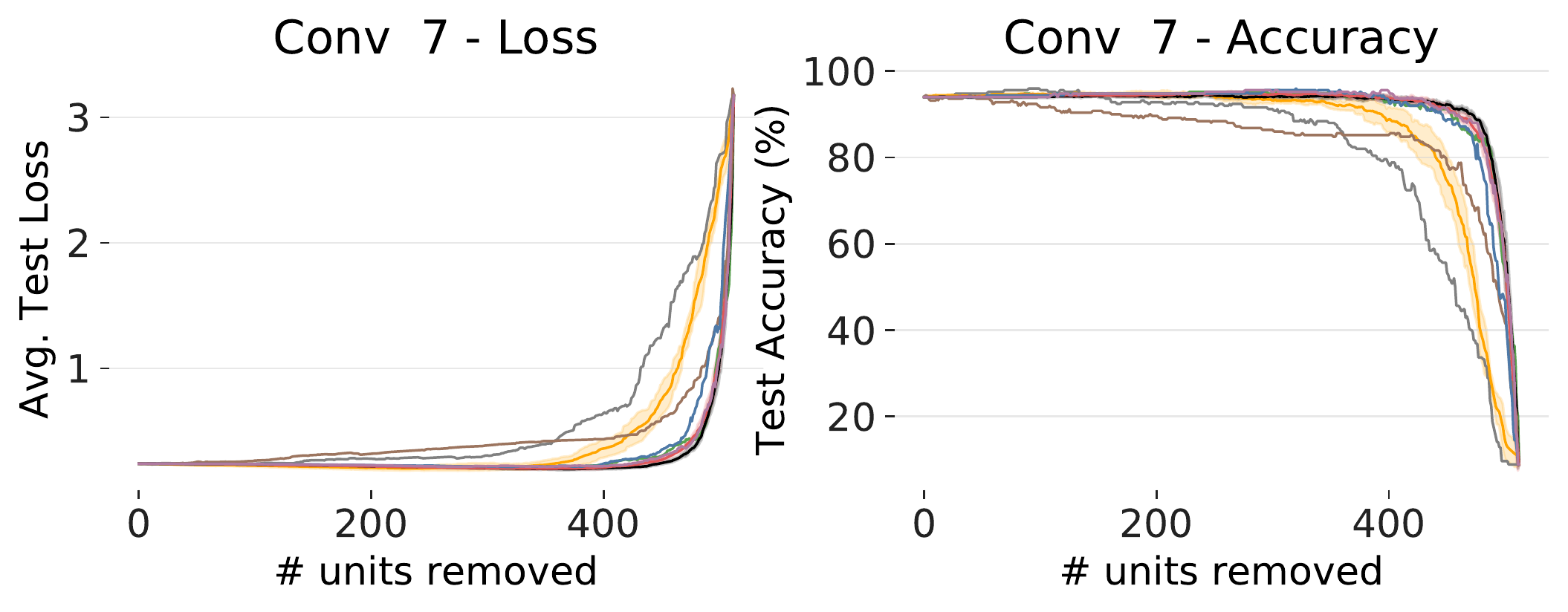}
\end{subfigure}
\begin{subfigure}
    \centering
    \includegraphics[width=0.45\textwidth]{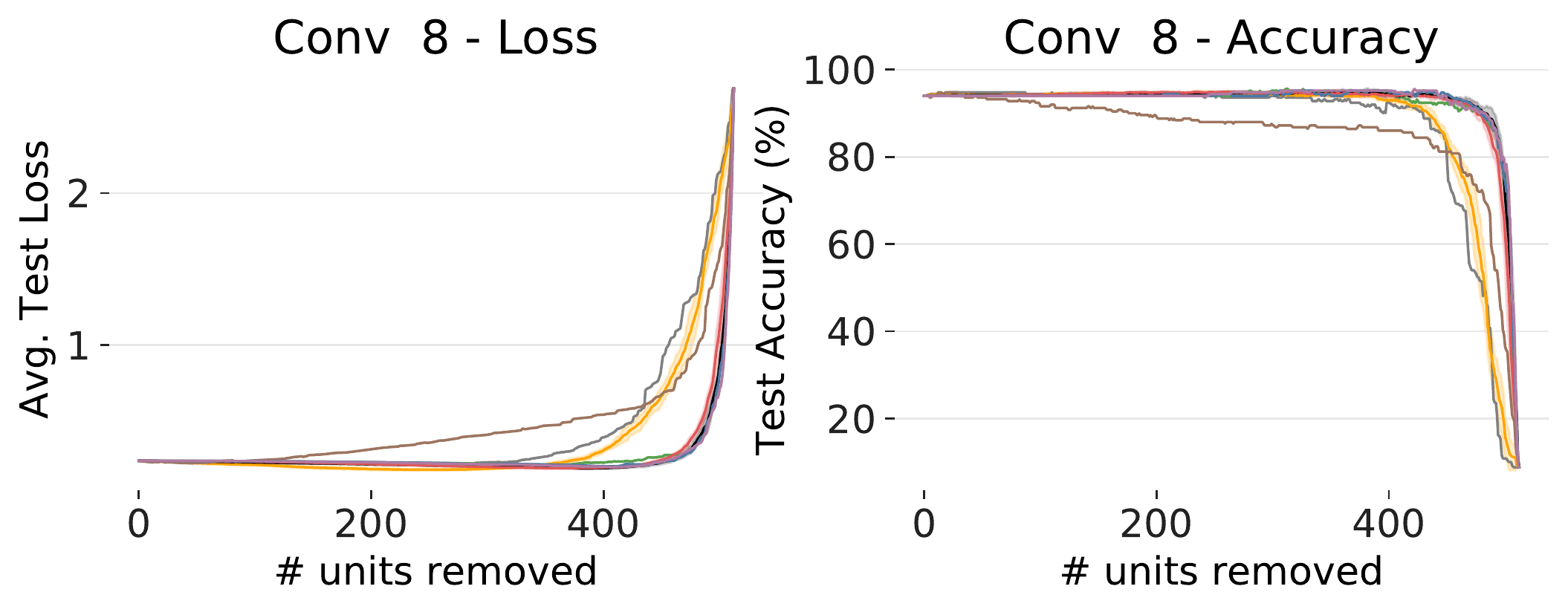}
\end{subfigure}
\begin{subfigure}
    \centering
    \includegraphics[width=0.45\textwidth]{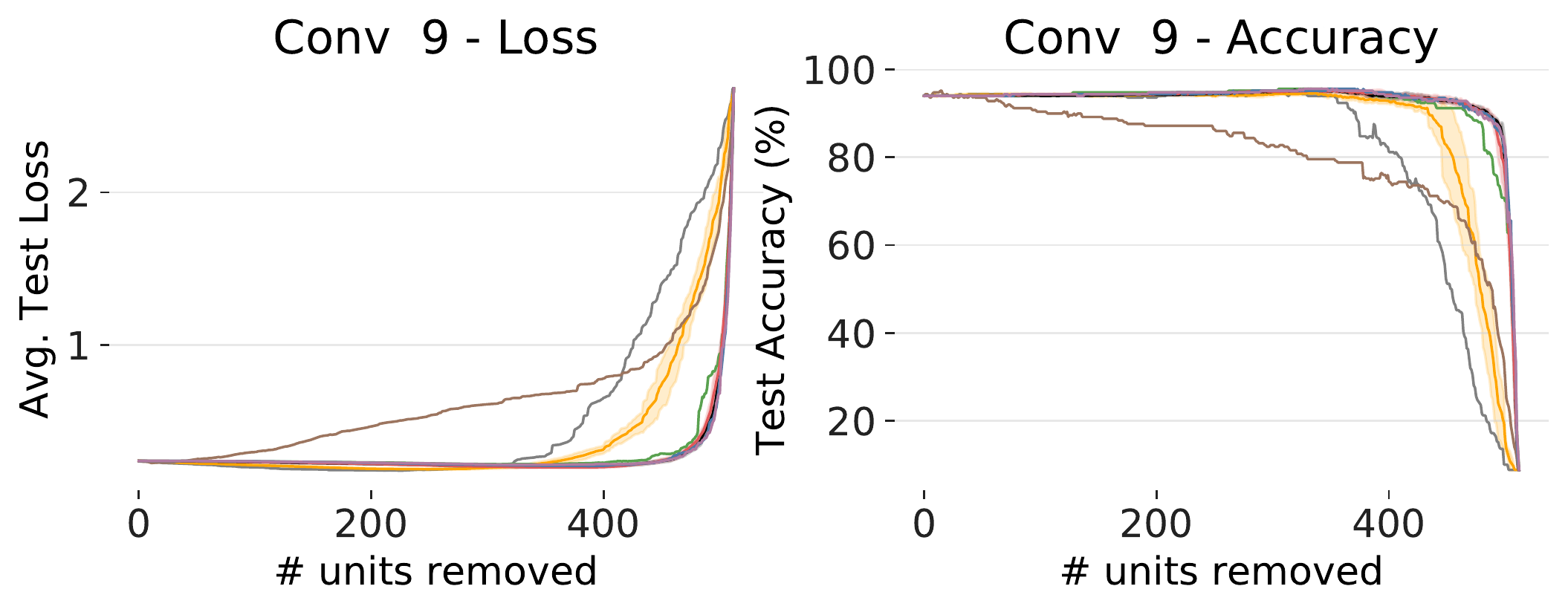}
\end{subfigure}
\begin{subfigure}
    \centering
    \includegraphics[width=0.45\textwidth]{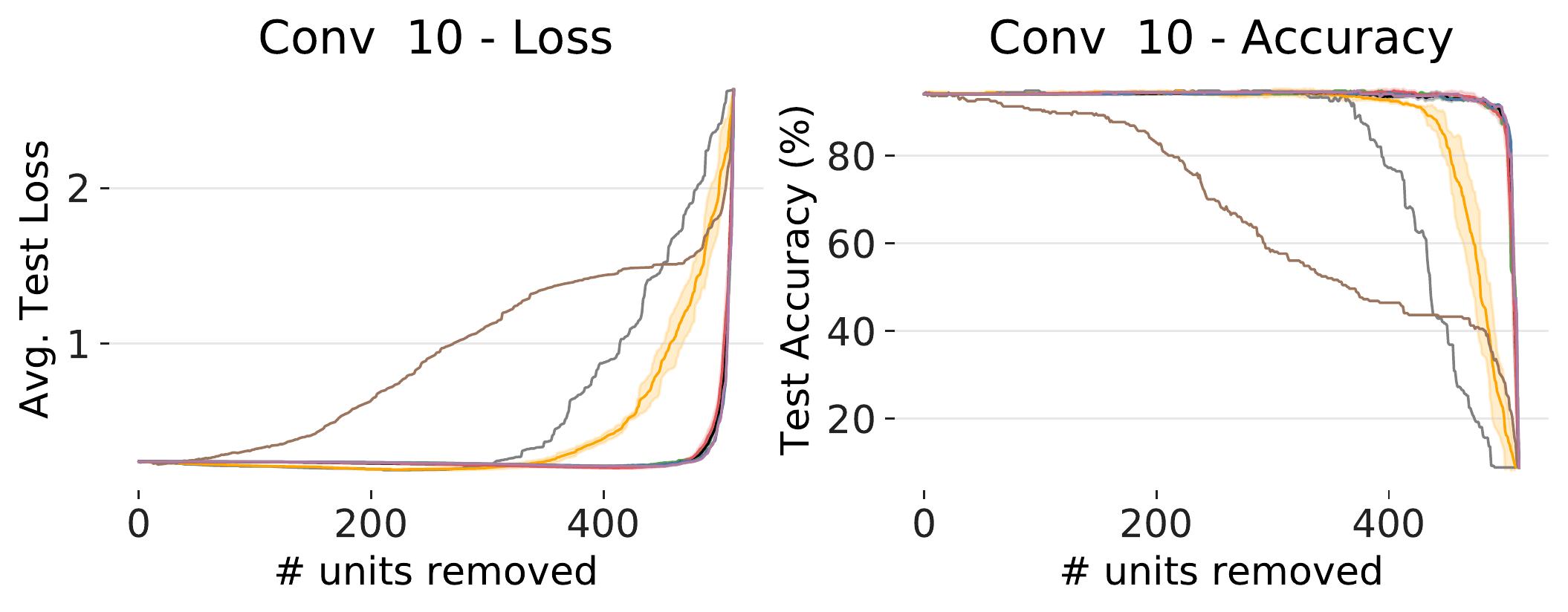}
\end{subfigure}%
\begin{subfigure}
    \centering
    \includegraphics[width=0.45\textwidth]{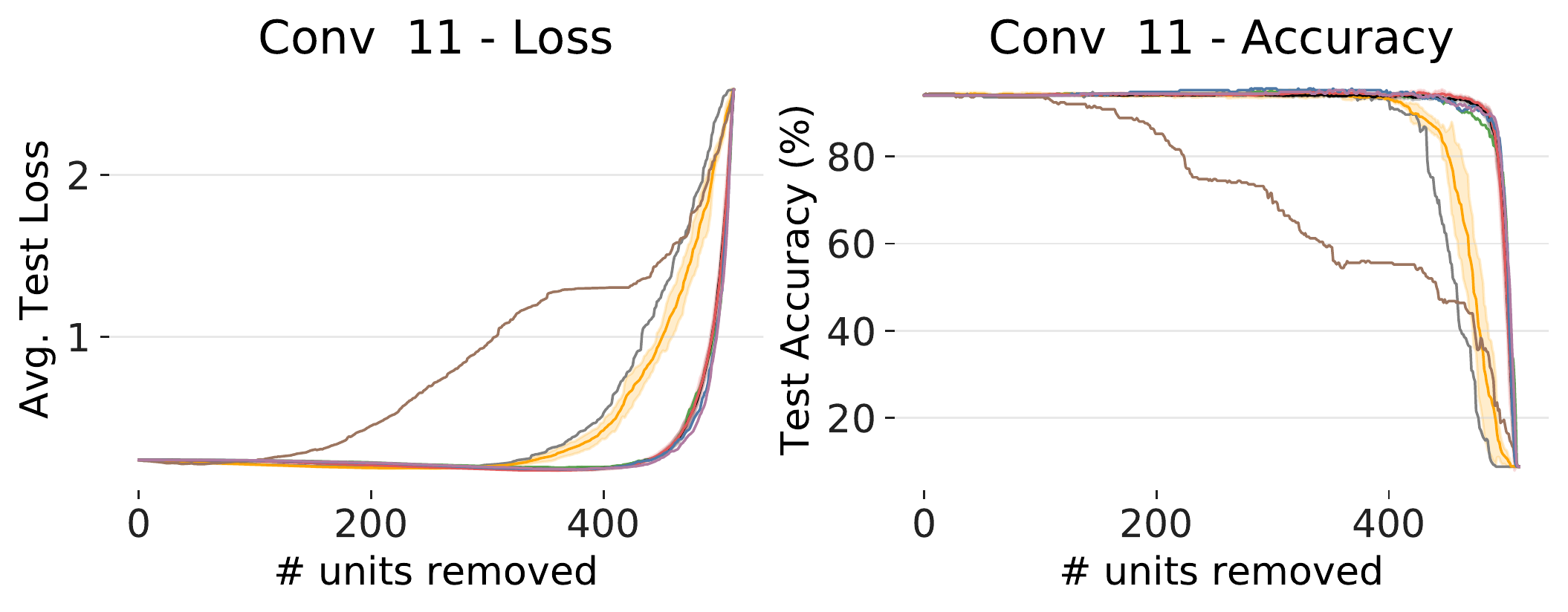}
\end{subfigure}
\begin{subfigure}
    \centering
    \includegraphics[width=0.45\textwidth]{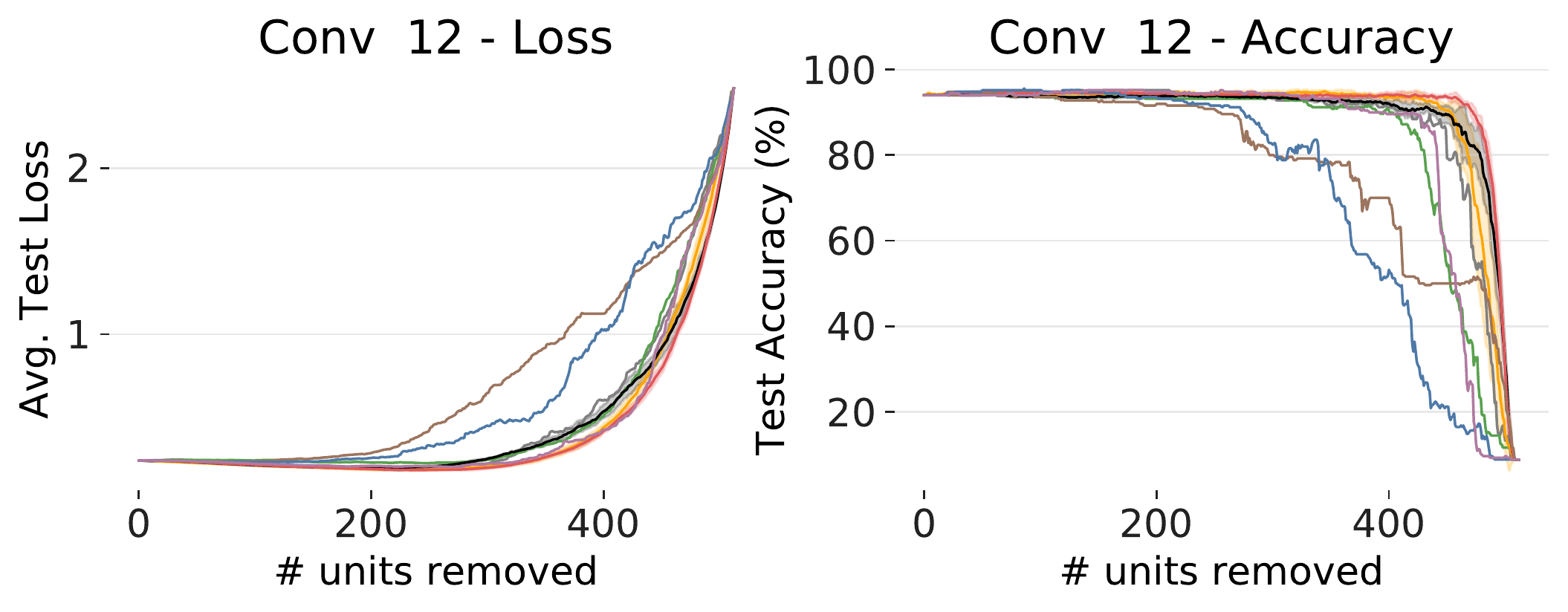}
\end{subfigure}
\begin{subfigure}
    \centering
    \includegraphics[width=0.45\textwidth]{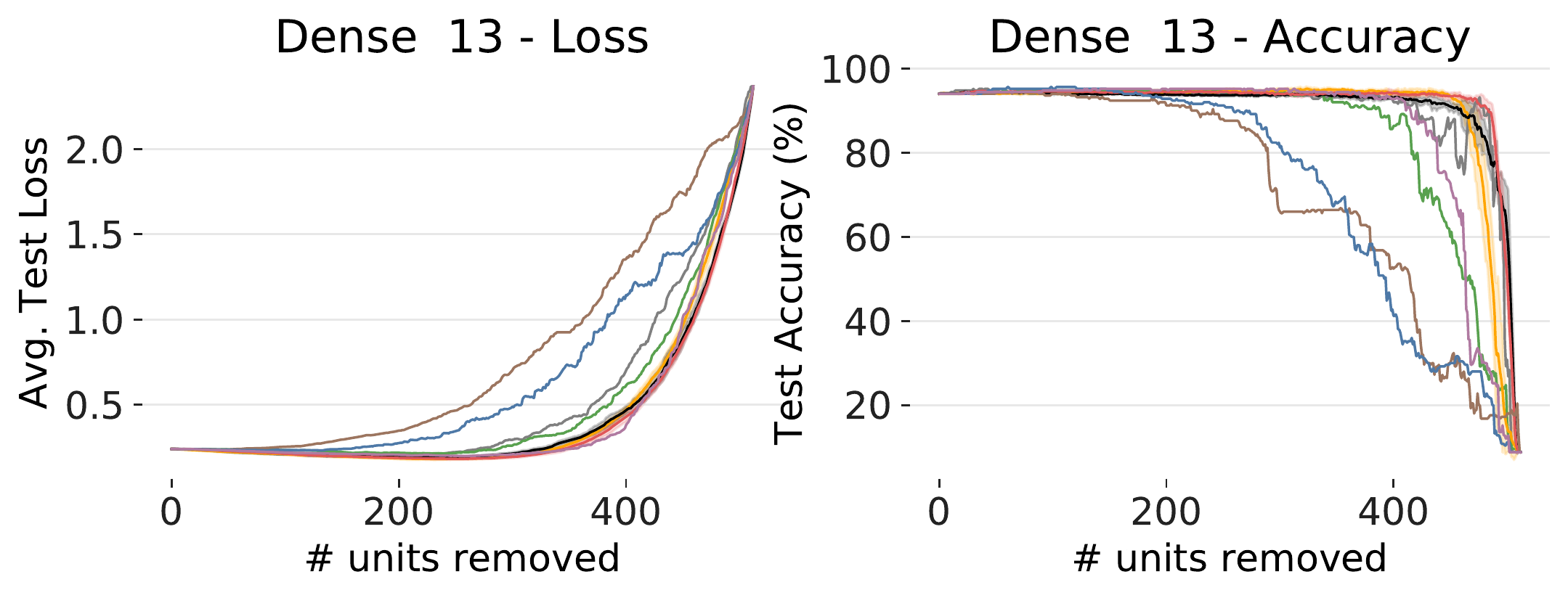}
\end{subfigure}%
\begin{subfigure}
    \centering
    \includegraphics[width=0.45\textwidth]{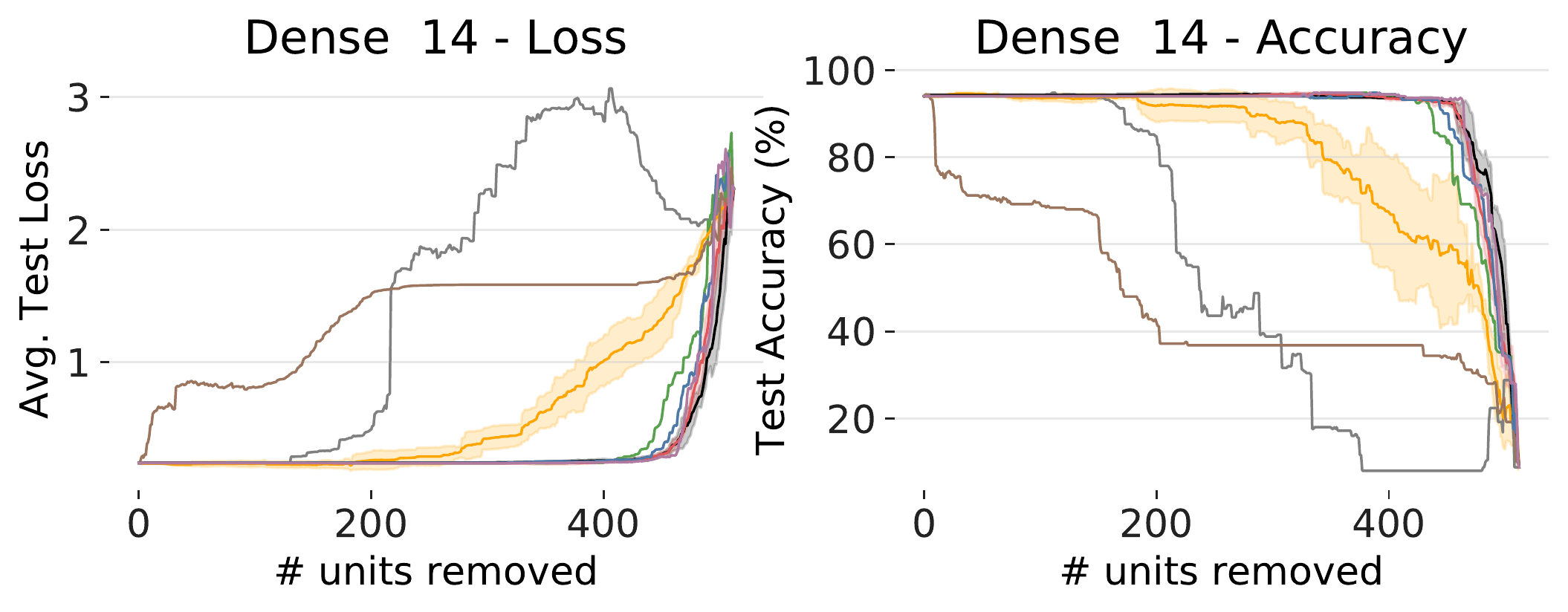}
\end{subfigure}

\begin{subfigure}
    \centering
    \includegraphics[width=0.9\textwidth]{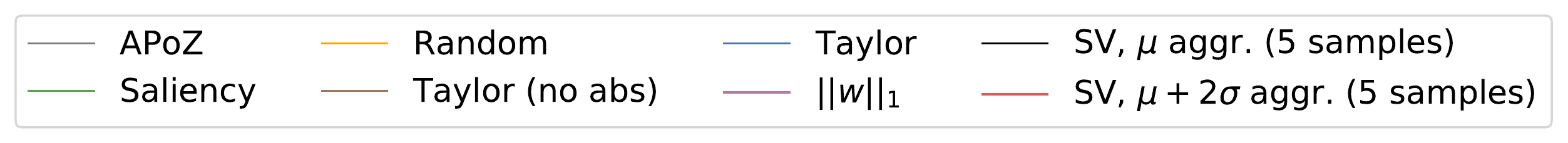}
\end{subfigure}
\caption{Test loss and accuracy on a VGG16 model on CIFAR-10, as the units of two of its layers are sequentially pruned.}
\end{figure}
\FloatBarrier

\section{Pruning with fine-tuning}

\begin{figure}[h!]
    \centering
    \includegraphics[width=\textwidth]{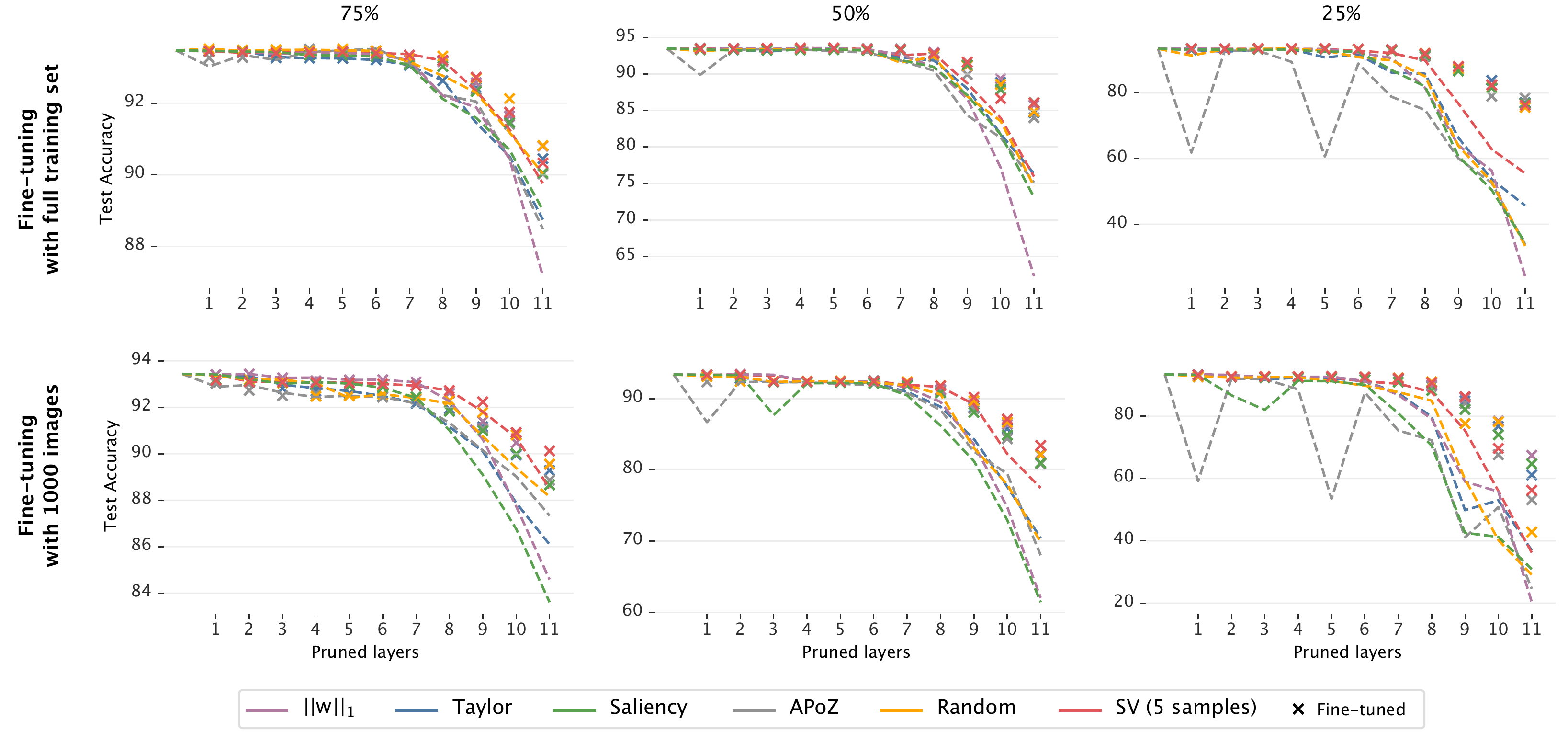}
    \caption{Performance of VGG16 on CIFAR-10 for different pruning metrics. For each pruned layer, we report the test accuracy before and after fine-tuning (dashed line and cross respectively). We fine-tune using the full-training data (top) and a small subset (bottom). Results averaged over 3 runs. Best seen in electronic form.}
    \label{fig:cifar_pruning_tune}
\end{figure}

Recently, it was showed that fine-tuning a network after pruning leads to the same recovery in performance regardless of the pruning criterion that had been used \cite{mittal2019studying}. The research showed this phenomenon under the assumption that the full training data is available for fine-tuning. 
In contrast, in our experimental section we showed that the Shapley value metric is superior in low-data regimes, i.e., when fine-tuning is not possible.
Figure \ref{fig:cifar_pruning_tune} shows  the performance of Shapley value pruning compared to other metrics \textit{with fine-tuning}. 

We consider the same VGG16 model pre-trained on CIFAR-10. In one experiment, we use the full training set and early stopping to run the fine-tuning after pruning each layer.
In a second experiment, we test fine-tuning with a smaller amount of data (but more than what used in the main experimental section of the paper).
We randomly take aside 1000 examples from the test set, since these have not been seen by the network during training, and we split them into two sets of equal size which will act as our new reduced training and validation sets. We keep the remaining examples in the original test set for the final performance evaluation. 
In both experiments, the fine-tuning after each pruning step is performed with SGD with fixed learning rate ($0.01$), no momentum and no weight decay. 

We notice that fine-tuning helps improving the performance of the network in almost all cases. More data allows retaining better performance after pruning, especially when we remove $75\%$ of the units.
While Shapley value pruning shows the best performance \textit{before} applying fine-tuning, the performance gap between different metrics \textit{after} fine-tuning is not statistically significant. In particular, we notice that depending on the amount of data, the pruning ratio, and the layer at which pruning is interrupted, the best performing metric varies significantly, although the average resulting accuracy is similar for all methods, including random pruning. These results are consistent with what found in \cite{mittal2019studying}.

\section{Details on the experimental setup}
We report here the details of the architectures used in our experiments.

\subsection{The role of sign for attributions}
We use MNIST \cite{lecun1998mnist} and CIFAR-10 \cite{krizhevsky2009learning} on a fully-connected network (2 hidden layers, 2048 units each and LeakyReLU\cite{maas2013leakyrelu} non-linearity  with 0.01 negative slope coefficient). The layers are initialized with Kaiming initialization \cite{he2015delving}. We estimate Shapley values with 5 sampling iterations on 10'000 images randomly taken from the training set. Shapley values are computed on the cross-entropy loss. The two hidden layers are pruned sequentially removing units with negative average Shapley value.

For the analysis in Figure \ref{fig:sv_ditribution} above, we train a custom architecture with 2 convolutional layers (32 and 64 filters), followed by one dense layer with 4096 units and a final linear layer which maps to 10 output classes.
The two convolutional layers are followed by Batch Normalization, ReLU activation and a $2\times2$ max-pooling while the hidden dense layer is followed by Batch Normalization and ReLU non-linearity. The network is trained for $50$ epochs on Fashion-MNIST \cite{xiao2017fmnist} using SDG with learning rate $0.01$ (halved every 15 epochs), momentum $0.9$ and weight decay $5*10^{-4}$. We also use random rotations and random horizontal flip as data augmentation during training. The final network reaches a test accuracy of $92.44\%$.

\subsection{Layer-wise pruning robustness}

The experiment with Fashion-MNIST used the network described in the previous subsection.

The experiment with CIFAR-10 uses a VGG16 network \cite{simonyan2014very} where the final dense layers have been replaced with smaller ones (512, 512, 10 output units). We use Batch Normalization and keep all other parameters as in the default PyTorch implementation. The network is trained for $160$ epochs, using SGD with learning rate $0.05$ (halved every 30 epochs), momentum $0.9$ and weight decay $5*10^{-4}$. We also use random cropping and random horizontal flip as data augmentation during training. The network has 15M parameters and reaches a test accuracy of $93.3\%$.

\subsection{Pruning in low-data regime}
For the experiments with CIFAR-10, we use the VGG16 model described in the previous subsection. 

For the experiment with CUB-200, we started from a VGG16 network pre-trained on ImageNet\footnote{We used the architecture and parameters provided by PyTorch Torchvision.}. We replaced the last dense layer with one randomly initialized having 200 output units instead of the original 1000,  resulting in a network with 135M parameters. Then, we fine-tuned the network end-to-end on the CUB-200 training set, using SGD with learning rate $0.001$, until reaching a test accuracy of $78\%$. 

\FloatBarrier
\bibliography{paper}
\bibliographystyle{abbrv}